\documentclass{article}

\usepackage[margin=1.17in]{geometry}

\usepackage{natbib}
\usepackage[utf8]{inputenc} 
\usepackage[T1]{fontenc}    
\usepackage{hyperref}       
\usepackage{url}            
\usepackage{booktabs}       
\usepackage{amsfonts}       
\usepackage{nicefrac}       
\usepackage{microtype}      
\usepackage{lipsum}
\usepackage{graphicx}
\usepackage{comment}
\graphicspath{ {./images/} }

\usepackage{authblk}
\usepackage{amsmath,amsfonts,bm}

\def\eqref#1{equation~\ref{#1}}

\def\1{\bm{1}}

\def\vh{{\bm{h}}}

\def\vv{{\bm{v}}}

\def\vx{{\bm{x}}}

\def\vxi{{\bm{\xi}}}

\def\mA{{\bm{A}}}
\def\mB{{\bm{B}}}
\def\mC{{\bm{C}}}

\def\mI{{\bm{I}}}
\def\mJ{{\bm{J}}}

\def\mQ{{\bm{Q}}}
\def\mR{{\bm{R}}}
\def\mS{{\bm{S}}}
\def\mT{{\bm{T}}}
\def\mU{{\bm{U}}}
\def\mV{{\bm{V}}}
\def\mW{{\bm{W}}}
\def\mX{{\bm{X}}}
\def\mY{{\bm{Y}}}

\def\mPhi{{\bm{\Phi}}}
\def\mLambda{{\bm{\Lambda}}}

\def\mGamma{{\bm{\Gamma}}}
\def\mPhi{{\bm{\Phi}}}
\def\mXi{{\bm{\Xi}}}

\DeclareMathAlphabet{\mathsfit}{\encodingdefault}{\sfdefault}{m}{sl}
\SetMathAlphabet{\mathsfit}{bold}{\encodingdefault}{\sfdefault}{bx}{n}

\DeclareMathOperator{\erf}{erf}

\title{A Dynamical Theory of LoRA in Continual Learning}

\author[1,${\dagger, \ddagger}$]{Théo Marchetta}
\author[1,${\dagger}$]{Filippo Alessandroni}
\author[2]{Alessandro Breccia}
\author[3]{Alessandro Ingrosso}
\author[1]{Federica Gerace}

\affil[1]{  Department of Mathematics, Alma Mater Studiorum – Università di
Bologna, Piazza di Porta San Donato 5, 40126 Bologna, Italy}
\affil[2]{Gatsby Computational Neuroscience Unit, University College London}
\affil[3]{Donders Centre for Neuroscience, Radboud University, Nijmegen, The Netherlands}
\date{}

\begin{document}
\maketitle
\noindent
$^\dagger$ Equal contributions.\\
$^\ddagger$ Correspondence to: \href{mailto:theo.marchetta@unibo.it}{theo.marchetta@unibo.it} 
\begin{abstract}

Despite the widespread use of Low-Rank Adaptation (LoRA), little is known about its dynamics in continual learning and the mechanisms by which low-rank updates affect catastrophic forgetting. We provide an asymptotically exact dynamical characterization of LoRA in a solvable two-task teacher-student model. In the high-dimensional online-learning limit, we derive a closed system of ordinary differential equations for a finite set of macroscopic order parameters, yielding exact expressions for the generalization errors throughout both the initial Task 1 learning phase and the subsequent LoRA fine-tuning on Task 2. The theory quantitatively matches finite-dimensional simulations and exposes two characteristic effects of LoRA: when the first-task representation is not yet specialized, low-rank adaptation reduces interference with features learned on the first task, but its initialization slows adaptation to the second task. Building on this mechanistic picture, we analyze a state-dependent masking strategy that freezes hidden units carrying the strongest first-task representations and restricts adaptation to the complementary subspace. This structural partitioning markedly reduces forgetting, while preserving plasticity on the new task. Our framework further clarifies the role of adapter rank: transfer improves only up to the intrinsic dimensionality of the target task and saturates beyond it, while forgetting continues to grow with
rank. These results provide a dynamical and geometric account of how low-rank adaptation organizes information across sequential tasks, and are qualitatively reproduced on sequential MNIST and CIFAR-100 using a shallow network and ResNet-18, respectively.
\end{abstract}

\section{Introduction}

Modern machine-learning systems are increasingly adapted to new tasks rather than trained from scratch. This paradigm is particularly important
for large pre-trained models, for which updating all parameters can be computationally and memory intensive. Parameter-efficient fine-tuning
(PEFT) methods address this problem by restricting adaptation to a small set of trainable parameters
\citep{xu2023parameterefficientfinetuningmethodspretrained,
han2024parameterefficientfinetuninglargemodels}. Among them, Low-Rank Adaptation (LoRA) \citep{hu2022lora} has become a widely used
approach: the pre-trained weights are frozen and adaptation is performed through a trainable low-rank perturbation.

Freezing the pre-trained weights, however, does not by itself guarantee that the behavior learned before fine-tuning is preserved. The LoRA update changes the effective representation seen by the network and can therefore interfere with features that were useful for previous tasks. This issue is particularly relevant in continual learning
\citep{Wang2024}, where models are trained sequentially and must acquire new information without catastrophically forgetting previously learned tasks \citep{McCloskey1989}. Recent methods have consequently sought to control the subspace in which LoRA updates occur, for example by constructing directions that reduce interference with previous tasks \citep{Liang2024} or by adapting underutilized spectral directions \citep{2604.01694}. Yet understanding \emph{why} low-rank adaptation retains or forgets information requires a dynamical description of how the adapter interacts with the representation learned before the task switch.

A growing theoretical literature has begun to characterize different aspects of LoRA, including its expressivity, convergence, initialization, and optimization dynamics
\citep{zeng2024the,xu2023parameterefficientfinetuningmethodspretrained,kim2025lora}.
Existing dynamical analyses either condition on a fixed pretrained state~\citep{nwemadji2026} or do not resolve the time dependence across the two stages~\citep{duranthon2026highdimensionaltheorylorafinetuning}.
These works provide important insights into LoRA adaptation, but leave open a complementary question central to continual learning: \emph{how does a low-rank update dynamically reorganize representations
learned on a previous task, and how does this geometry jointly shape transfer to the new task and forgetting of the old one?}

We address this question in a solvable two-task teacher-student model, building on the high-dimensional online-learning framework of
\citet{lee2021continual,lee2022}. Task 1 is learned by standard SGD; at the task switch, the learned representation is frozen and Task 2 is learned through the LoRA update. In the high-dimensional limit, we derive a closed system of deterministic ordinary differential equations (ODEs) for a finite set of macroscopic overlaps that determine the generalization errors on both tasks. LoRA changes the dynamical equations after the switch: the pretrained overlaps become fixed, while additional order parameters track the geometry of the LoRA adapter relative to the frozen representation
and to both teachers. This framework reveals reduced interference but slower adaptation under LoRA in the unspecialized regime, and allows us to study subspace
restriction, adapter rank, and task similarity within a common stability-plasticity picture.

\paragraph{Main contributions.}
\begin{itemize}
    \item \textbf{Dynamical theory of sequential LoRA.} 
    We derive an asymptotically exact high-dimensional description of Task 1 feature learning followed by Task 2 LoRA adaptation. The stochastic dynamics close onto a finite system of ODEs that jointly determine transfer, forgetting, and representation geometry.

    \item \textbf{LoRA-specific mechanisms and structured adaptation.}
    We show that LoRA freezes the pretrained overlaps while introducing new adapter-representation and adapter-teacher overlaps, revealing, as long as the Task 1 representation is not specialized, reduced interference but slower adaptation from the initialization \citep{hayou2024}. Motivated by existing subspace-constrained adaptation methods \citep{Liang2024,2604.01694}, we incorporate a state-dependent masking
    rule into the same dynamical theory, showing that it strongly reduces forgetting while preserving Task 2 performance; we observe the same qualitative behavior on sequential MNIST and Split CIFAR-100 using a shallow network and a deep convolutional network (ResNet-18), respectively.

    \item \textbf{Rank, similarity, and stability-plasticity.}
    We characterize how adapter rank and task similarity control transfer and forgetting: useful transfer saturates once the adapter can represent the target feature space, while forgetting increases. In the unspecialized regime, interference peaks at intermediate task similarity while, in both regimes, structured masking strongly suppresses it.
\end{itemize}

The code used in the present manuscript is provided in \href{https://github.com/thmarchetta/online_learning_LoRA}{this  repository}.

\section{Related Work}

\paragraph{High-dimensional learning dynamics.}
Our analysis builds on a rigorous statistical-physics description of online learning \citep{Goldt2019}, where high-dimensional stochastic updates reduce to
deterministic dynamics for a finite set of macroscopic order parameters
\citep{gardner1989,PhysRevE.52.4225,Saad1995,biehl1999}. The generalization error can then be expressed in terms of these sufficient statistics across training time. \citet{lee2021continual} extended this formalism to continual learning under standard SGD, showing how task similarity controls transfer and forgetting; subsequent work studied feature re-use and optimal control
\citep{lee2022,mori2025optimal}. Our Task 1 dynamics follow this framework, but after the switch we freeze the learned representation and optimize a factorized low-rank perturbation, which requires a different
macroscopic closure.

\paragraph{Continual learning with LoRA.}
Continual-learning methods mitigate catastrophic forgetting through regularization, replay, or architectural separation \citep{doi:10.1073/pnas.1611835114,1902.10486,
rusu2022progressiveneuralnetworks}. For low-rank adaptation, empirical comparisons with standard fine-tuning show that LoRA tends to learn less and forget less \citep{biderman2024lora}, but that this is not systematic: LoRA can forget more when fine-tuning is repeated over sequences of tasks \citep{shuttleworth2025}. Rather than relying on the low rank alone, recent LoRA-based approaches constrain the adapter update \citep{lu-etal-2025-controlled,Wei2025,Liang2024,Che_2026}. Most relevant here, InfLoRA selects directions designed to reduce previous-task interference \citep{Liang2024}, while MiCA restricts adaptation using the spectral structure of the pretrained model \citep{2604.01694}. Our objective is complementary: rather than proposing a new subspace-selection principle, we use a solvable dynamical model to analyze how constraining the LoRA update relative to previously learned features affects the stability-plasticity trade-off.

\paragraph{Theoretical analyses of LoRA.}
LoRA theory has addressed expressivity, optimization, convergence, and
generalization \citep{zeng2024the,pmlr-v258-xu25h,kim2025lora,kratsios2025}. Most closely related, \citet{nwemadji2026} study LoRA fine-tuning dynamics conditional on a pretrained state. Our setting instead tracks the full sequential process: we dynamically generate the pretrained state through Task 1 learning and subsequently track both Task 2 acquisition and Task 1 forgetting. \citet{duranthon2026highdimensionaltheorylorafinetuning} instead provide a high-dimensional asymptotic theory of pre-training and LoRA fine-tuning in a solvable attention model, without resolving the time-dependent
training dynamics across the two stages. Our framework therefore connects dynamical continual-learning theory with a time-resolved theory of low-rank adaptation.

\section{Continual Learning with LoRA: Problem Setting}
\label{sec:LoRA_problem_setting}

We study continual learning in a two-task teacher-student setting \citep{gardner1989,lee2021continual}. Let

\begin{equation}
\phi(\boldsymbol{\xi};\mJ, \mathbf v)
=
\sum_{i=1}^{D}
v_i\,
g\!\left(
\frac{\mJ_i\boldsymbol{\xi}}{\sqrt N}
\right)
\end{equation}

denote a two-layer fully connected network with $D$ hidden units, first-layer weights $\mJ\in\mathbb{R}^{D\times N}$, readout weights $\boldsymbol v\in\mathbb{R}^{D}$, and activation function $g$. The notation $f(x ;\, y)$ distinguishes the variable $x$ from the fixed parameter $y$. Inputs are sampled
independently as 
$\boldsymbol{\xi}\sim\mathcal{N}(\mathbf{0},\mI_N).$

\paragraph{Teachers and task similarity.}
The two tasks are generated by fixed teacher networks indexed by
$\ast\in\{\dagger,\ddagger\}$, with parameters
$(\mW^\ast,\boldsymbol v^\ast)$, where
$\mW^\ast\in\mathbb{R}^{M\times N}$ and
$\boldsymbol v^\ast\in\mathbb{R}^{M}$. Their targets are

\begin{equation}
y^\ast=\phi(\boldsymbol{\xi};\mW^\ast,\boldsymbol v^\ast).
\end{equation}

Teacher $\dagger$ defines Task 1 and teacher $\ddagger$ Task 2. To control task similarity, we draw $W^\dagger_{ij}\overset{\mathrm{iid}}{\sim}\mathcal{N}(0,1)$ and set $
\mW^\ddagger
=
c\,\mW^\dagger+\sqrt{1-c^2}\,\mathbf{Z},
\
\mathbf{Z}_{ij}\overset{\mathrm{iid}}{\sim}\mathcal{N}(0,1)
$, with $\mathbf{Z}$ independent of $\mW^\dagger$ and $c\in[0,1]$. In the high-dimensional limit $N\to\infty$ with $M=O(1)$,

\begin{equation}
\frac{1}{N}\mW^\dagger(\mW^\ddagger)^T
\longrightarrow c\,\mI_M.
\end{equation}

Hence, $c=0$ corresponds to asymptotically orthogonal teacher features, whereas $c=1$ gives identical first-layer teacher representations. For both teachers, we take uniform readout weights with opposite signs, $\vv_i^\dagger = +1$ and $\vv_i^\ddagger = -1$ for $i\in[M]$.

\paragraph{Student and sequential learning protocol.}

The student has shared first-layer weights
$\mJ\in\mathbb{R}^{K\times N}$ and task-specific readout heads $\boldsymbol h^\dagger,\boldsymbol h^\ddagger\in\mathbb{R}^{K}$, with prediction

\begin{equation}
\hat y^\ast
=
\phi(\boldsymbol{\xi};\mJ,\boldsymbol h^\ast).
\end{equation}

Task identity is therefore known at evaluation time, and since $\vh^\dagger$ is frozen after the switch (below), the forgetting we measure is caused by changes in the shared first layer alone. Unless stated otherwise, we consider $K=2M$, so that the student has sufficient hidden-layer capacity to represent both teachers without an intrinsic width bottleneck.

Training proceeds sequentially. During Task 1, $\mJ$ and $\boldsymbol h^\dagger$ are optimized by online SGD on the squared loss using
samples generated by teacher $\dagger$. Writing $\mu$ for the number of examples presented, the macroscopic dynamics evolve on the rescaled time $\tau = \frac{\mu}{N}$, and each task is trained for $\tau \in [0, \alpha]$; $\alpha$ therefore sets the number of training steps per task, $T = \alpha N$ (see Appendix \ref{app:theory_task1}). At the task switch, examples begin to be generated by teacher $\ddagger$. Let $\mJ_s$ denote the first-layer weights at the task switch. During Task 2, $\mJ_s$ and $\boldsymbol h^\dagger$ are frozen, while adaptation is performed through a LoRA update

\begin{equation}
\label{eq:LoRA_adapter}
\mJ
=
\mJ_s+\Delta\mJ,
\qquad
\Delta\mJ
=
\frac{\gamma}{\sqrt L}\mB\mA,
\end{equation}

where $\mA\in\mathbb{R}^{L\times N}$,
$\mB\in\mathbb{R}^{K\times L}$,
$L$ is the adapter rank, and $\gamma$ controls the update scale. During Task 2, $\mA$, $\mB$, and $\boldsymbol h^\ddagger$ are trainable. The LoRA factors are initialized so that $\Delta\mJ=\mathbf{0}$ at the task switch, ensuring that inserting the adapter does not immediately perturb the Task 1 representation. Details on initialization of the low-rank adapters are reported in Appendix~\ref{app:LoRA}.

\paragraph{Generalization error and high-dimensional limit.}
Performance on task $\ast\in\{\dagger,\ddagger\}$ is measured by the population generalization error
\begin{equation}
\epsilon^\ast
=
\frac12
\mathbb{E}_{\boldsymbol{\xi}}
\biggl[ \Bigl(
\phi(\boldsymbol{\xi};\mJ,\boldsymbol h^\ast)
-
\phi(\boldsymbol{\xi};\mW^\ast,\boldsymbol v^\ast)
\Bigr)^2 \biggr].\label{eq:test-error}
\end{equation}
We work in the online-learning regime, where each SGD step uses an independent sample from the data-generating distribution.
Throughout we take $\eta_\mJ$ ,$\eta_\mA$, $\eta_\mB$ and $\eta_\vh$ for the learning rates of the first layer, the two adapter factors and the readouts; their $N$-scalings are fixed in Appendix~\ref{app:theory_Task2} and their values are reported in Appendix~\ref{app:hyperparameters}.
In the following, we show that, in the limit
$N\to\infty$ with $K,M,L=O(1)$, the stochastic training dynamics concentrate onto a closed deterministic system for a finite set of macroscopic order parameters, from which we can track the evolution of the generalization error of both tasks with training time.

\section{A Dynamical Theory of LoRA in Continual Learning}
\label{sec:theory}

\begin{figure}[t!]
    \centering
    \begin{minipage}[c]
    {0.43\linewidth}
    {\raggedright \textbf{a)}\par}
        \vspace{2pt}
        \centering
        \includegraphics[
            width=\linewidth]{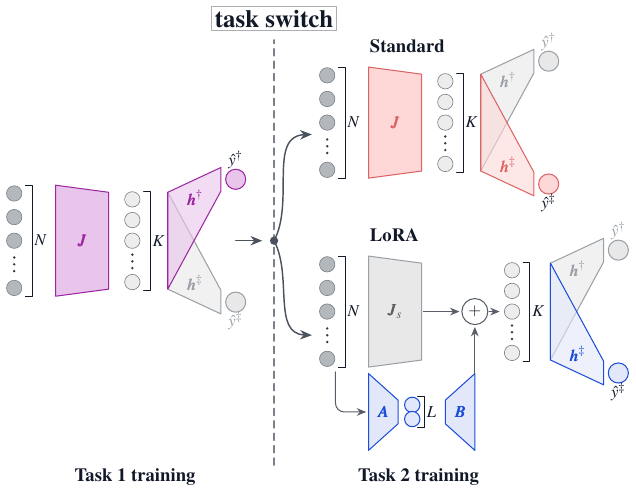}
    \end{minipage}
    \hfill
    \begin{minipage}[c]{0.55\linewidth}
    {\raggedright \textbf{b)}\par}
        \vspace{2pt}
        \centering
        \includegraphics[
            width=\linewidth
        ]{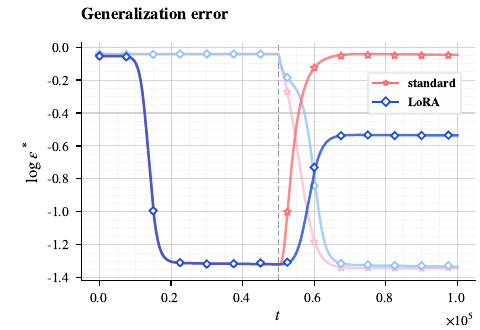}
    \end{minipage}
    \vspace{2mm}

    \begin{minipage}[t]{0.48\linewidth}
    {\raggedright \textbf{c)}\par}
        \vspace{2pt}
        \centering
        \includegraphics[
            width=\linewidth
        ]{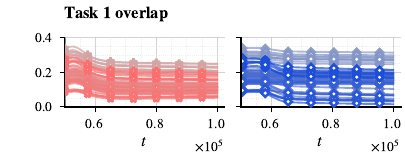}
    \end{minipage}
    \hfill
    \begin{minipage}[t]{0.48\linewidth}
    {\raggedright \textbf{d)}\par}
        \vspace{2pt}
        \centering
        \includegraphics[
            width=\linewidth
        ]{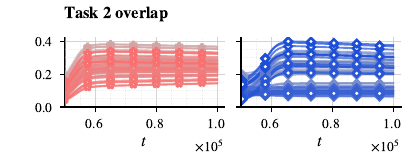}
    \end{minipage}

    \caption{\textbf{Generalization and representation dynamics under LoRA and full fine-tuning for a sequential training.} \textit{a)} Schematic of the sequential training setting. We first train the first-layer weights and a readout head on a synthetic dataset corresponding to Task 1. We then train on a correlated dataset corresponding to Task 2, where we either update the first-layer weights or freeze them and instead train a low-rank adapter. In both cases, we train a second readout head. \textit{b)} Task 1 (dark) and Task 2 (light) generalization errors for full fine-tuning and LoRA. The gray dashed line marks the task switch. \textit{c--d)} Student overlaps with teachers as defined in Eqs. (\ref{eq:tilde_x_rho})-(\ref{eq:tilde_x_nu}) after the task switch for both full fine-tuning (red) and LoRA (blue). Solid lines denote theoretical ODE predictions and markers finite-dimensional simulations. Parameters: $N=10^3$, $K=10$, $M=5$, $L=5$, $c=0.5$, $\alpha=50$.}
    \label{fig1:StandardVSLoRA}
\end{figure}

We now derive a closed macroscopic description of the sequential learning dynamics in the high-dimensional limit for activation function $g(z)=\mbox{erf}(z/\sqrt{2})$. Our analysis builds on the online teacher-student framework of \citet{lee2021continual,lee2022}. The key difference arises after the task switch: instead of continuing to update the student first-layer weights, we freeze the Task 1 representation and optimize a factorized
low-rank perturbation. This changes both the microscopic dynamics and the set of macroscopic quantities required to obtain a closed theory.

\paragraph{Task 1: standard feature learning.}
During training on Task 1, the student weights $(\mJ^\mu, (\boldsymbol{h}^\dagger)^\mu)$ are updated via online SGD on the squared loss. The resulting prediction error on the $\mu$-th example is
\begin{equation}
\Delta^{\dagger,\mu}
=
\sum_{k=1}^{K}
(h_k^\dagger)^\mu g(x_k^\mu)
-
\sum_{m=1}^{M}
v_m^\dagger g(\rho_m^\mu).
\end{equation}

where, for an input $\boldsymbol{\xi}^{\mu}$, $\rho_m^\mu$ and $x_k^\mu$ define the Task 1 teacher and student preactivations
\begin{equation}
\rho_m^\mu
=
\frac{\mW_m^\dagger\boldsymbol{\xi}^{\mu}}{\sqrt N},
\qquad
x_k^\mu
=
\frac{\mJ_k^\mu\boldsymbol{\xi}^{\mu}}{\sqrt N}.
\end{equation}

The scaling in $N$ is chosen such that the macroscopic quantities evolve on the time scale $\tau=\mu/N$ as $N \rightarrow \infty$, this phase coincides with the standard continual-learning dynamics of \citet{lee2021continual}. All details on these existing results can be found in Appendix~\ref{app:theory_task1}.

\paragraph{Task 2: low-rank adaptation of a frozen representation.}
At the task switch, the feature matrix is frozen at $\mJ_s$ and the
effective representation becomes

\begin{equation}
\mJ
=
\mJ_s+\frac{\gamma}{\sqrt L}\mB\mA.
\end{equation}

During this phase, $\mA$, $\mB$, and $\boldsymbol h^\ddagger$ are updated by
online SGD, while $\mJ_s$ and $\boldsymbol h^\dagger$ remain frozen. The resulting Task 2 prediction error is therefore
\begin{equation}
\Delta^\ddagger
=
\sum_{k=1}^{K}
h_k^\ddagger g(\tilde x_k)
-
\sum_{p=1}^{M}
v_p^\ddagger g(\nu_p).
\end{equation}

which depends on the Task-2 teacher fields, the frozen student fields and the $L$ LoRA fields
\begin{equation}
\nu_p^\mu
=
\frac{\mW_p^\ddagger\boldsymbol{\xi}^{\mu}}{\sqrt N},
\qquad
x_k
=
\frac{(\mJ_s)_k\boldsymbol{\xi}}{\sqrt N},
\qquad
z_i
=
\frac{\mA_i\boldsymbol{\xi}}{\sqrt N},
\end{equation}

so that the adapted student preactivation is
\begin{equation}
\tilde x_k
=
x_k+
\frac{\gamma}{\sqrt L}
\sum_{i=1}^{L} \mB_{ki}z_i .
\end{equation}

\paragraph{Macroscopic order parameters.}
The population generalization errors in (\ref{eq:test-error}) depend on the $N$-dimensional input only through the scalar fields. Since
$\boldsymbol{\xi}\sim\mathcal{N}(0,\mI_N)$ and all fields above are linear functions of $\boldsymbol{\xi}$, they are jointly zero-mean
Gaussian. Their distribution is therefore fully determined by their second moments, which are normalized inner products between the corresponding weight vectors. Before the task switch, we track

\begin{equation}
\mQ_{kl}
:=
\langle x_kx_l\rangle,
\qquad
\mR_{km}
:=
\langle x_k\rho_m\rangle,
\qquad
\mU_{kp}
:=
\langle x_k\nu_p\rangle,
\end{equation}

together with the fixed teacher overlaps

\begin{equation}
\mT_{mn}
:=
\langle\rho_m\rho_n\rangle,
\qquad
\mV_{mp}
:=
\langle\rho_m\nu_p\rangle,
\qquad
\mS_{pq}
:=
\langle\nu_p\nu_q\rangle,
\end{equation}

which, by the teacher construction satisfy $\mT=\mS=\bm I_M$ and
$\mV=c\,\bm I_M$ in the high-dimensional limit.

Equivalently, by replacing the field definition

\begin{equation}
\mQ_{kl}
=
\frac{1}{N}\mJ_k\mJ_l^T,
\qquad
\mR_{km}
=
\frac{1}{N}\mJ_k(\mW_m^\dagger)^T,
\qquad
\mU_{kp}
=
\frac{1}{N}\mJ_k(\mW_p^\ddagger)^T,
\end{equation}

with analogous expressions for $\mT,\mV,\mS$.
Thus $\mQ$ describes the geometry of the student representation, $\mR$ and $\mU$ its alignment with Tasks~1 and~2, and $\mT,\mV,\mS$ the fixed geometry of the teachers.

LoRA introduces four additional overlap matrices,

\begin{equation}
\mPhi_{ij}
:=
\langle z_i z_j\rangle,
\qquad
\mXi_{ki}
:=
\langle x_k z_i\rangle,
\qquad
\mLambda_{mi}
:=
\langle \rho_m z_i\rangle,
\qquad
\mGamma_{pi}
:=
\langle \nu_p z_i\rangle.
\end{equation}

In weight space,
\begin{equation}
\mPhi_{ij}
=
\frac{1}{N}\mA_i\mA_j^T,
\qquad
\mXi_{ki}
=
\frac{1}{N}(\mJ_s)_k\mA_i^T,
\qquad
\mLambda_{mi}
=
\frac{1}{N}\mW^\dagger_m\mA_i^T,
\qquad
\mGamma_{pi}
=
\frac{1}{N}\mW^\ddagger_p\mA_i^T.
\end{equation}

Hence $\mPhi$ describes the geometry of the trainable LoRA directions, $\mXi$ their alignment with the frozen student
representation, and $\mGamma$ and $\mLambda$ their alignment with the Task 2 and Task 1 teachers, respectively. Since $\mB\in\mathbb{R}^{K\times L}$ remains finite-dimensional as $N\to\infty$, its entries are tracked explicitly.

\paragraph{How LoRA changes the dynamical closure.}
This is the central modification relative to standard continual learning. Under full fine-tuning, the student overlaps themselves
continue to evolve after the task switch. Under LoRA, $\mQ_s,\mR_s,\mU_s$ are fixed at their Task 1 values, and the geometry of the effective representation is reconstructed from these frozen overlaps and the adapter variables. In particular,
\begin{align}
      & \langle \tilde x_k\rho_m\rangle
=
(\mR_s)_{km}
+
\frac{\gamma}{\sqrt L}
(\mB\mLambda^T)_{km},
\label{eq:tilde_x_rho}\\
    &\langle \tilde x_k\nu_p\rangle
    =
    (\bm U_s)_{kp}
    +
    \frac{\gamma}{\sqrt L}
    (\mB\mGamma^T)_{kp}, \label{eq:tilde_x_nu}
\end{align}
and
\begin{equation}
\begin{aligned}
\langle \tilde x_k\tilde x_l\rangle
&=
(\mQ_s)_{kl}
+
\frac{\gamma}{\sqrt L}
\left[
(\mXi\mB^T)_{kl}
+
(\mB\mXi^T)_{kl}
\right] 
+
\frac{\gamma^2}{L}
(\mB\mPhi\mB^T)_{kl}.
\end{aligned}
\end{equation}

These relations completely determine the covariance matrix of the adapted preactivations and therefore the population errors on both tasks. More precisely, in the high-dimensional limit, during Task 2 training the generalization errors of both tasks are functions of the order parameters
\begin{equation}
    \begin{split}
        \lim_{N \rightarrow \infty} \epsilon^\dagger
    &=
    \epsilon^\dagger
    (
    \mXi,\mPhi,\mLambda,
    \mB \, ; \,\,
    \mQ_s,\mR_s,\mT, \boldsymbol h^\dagger,
    \boldsymbol v^\dagger
    )\\
    \lim_{N \rightarrow \infty} \epsilon^\ddagger
    &=
    \epsilon^\ddagger
    (
    \mXi,\mPhi,\mGamma,
    \mB , \vh^\ddagger\, ; \,\,
    \mQ_s,\mU_s,\mS,
    \vv^\ddagger
    )
    \end{split}
\end{equation}

The corresponding explicit expressions are given in Appendix~\ref{app:theory_Task2}.

\paragraph{Deterministic high-dimensional dynamics.}
The evolution equations follow by combining the microscopic SGD updates with the definitions above and taking the limit $N\to\infty$ at fixed $\tau=\mu/N$. As an illustration, consider
\begin{equation}
\mGamma_{pi}
=
\frac{1}{N}
\mW_p^\ddagger\mA_i^T,
\end{equation}

which measures the alignment between the $i$-th LoRA direction and the $p$-th Task 2 teacher feature. Its evolution is

\begin{equation}
\frac{\mathrm{d}\mGamma_{pi}}{\mathrm{d}\tau}
=
-\eta_\mA\frac{\gamma}{\sqrt L}
\sum_{k=1}^{K}
\vh_k^\ddagger \mB_{ki}
\left\langle
\Delta^\ddagger
g'(\tilde x_k)\nu_p
\right\rangle .
\end{equation}

Analogous calculations yield a closed deterministic system for $\mPhi,\mXi,\mGamma,\mLambda,\mB$ and
$\boldsymbol h^\ddagger$. For $g(z)=\mbox{erf}(z/\sqrt{2})$, all Gaussian expectations can be evaluated in
closed form as algebraic functions of the instantaneous order parameters. The complete ODE system and the corresponding generalization-error expressions for Task 2 training are provided in Appendix~\ref{app:theory_Task2}.


\paragraph{LoRA versus full fine-tuning.}

Figure~\ref{fig1:StandardVSLoRA}b compares the resulting LoRA dynamics (blue) with standard full fine-tuning (red). Full fine-tuning rapidly increases the Task 1 error after the switch, whereas LoRA preserves substantially more of the previously learned representation while reaching a comparable asymptotic Task 2 error. The theoretical trajectories (solid lines) closely match finite-dimensional simulations (markers).

The overlap dynamics provide a geometric explanation. Under full fine-tuning, the shared representation itself moves toward Task 2, thereby modifying features acquired on Task 1. Under LoRA, the pretrained component remains fixed and the change in Task 1 and Task 2 alignment is mediated only by $\mB\mLambda^T$ and $\mB\mGamma^T$, respectively. As illustrated in Fig.~\ref{fig1:StandardVSLoRA}c-d, LoRA increases alignment with the new task while preserving a larger fraction of the Task 1 alignment than full fine-tuning. This provides a direct representation-level explanation for its reduced forgetting.

In Fig.~\ref{fig1:StandardVSLoRA}, the task switch occurs before the student fully specializes to the Task 1 teacher, as indicated by Fig.~\ref{fig1:StandardVSLoRA}c. This regime is practically relevant, since training is typically stopped once a target performance is reached rather than after complete representational specialization; in addition, the time constant for symmetric-subspace escape grows with $K$, so full specialization at $K > M$ requires substantially longer training. The reduced interference of LoRA is specific to the unspecialized regime: once the student has specialized to the Task 1 teacher, LoRA re-uses the specialized units and forgets as much as full fine-tuning. Results in the fully specialized regime are reported in Appendix~\ref{app:specialization}.

LoRA also adapts more slowly at early times, as observed empirically~\citep{2402.09353,pmlr-v267-li25bm}. Our theory attributes this to the dynamics of the adapter: with $\bm A$ initialized at zero, the Task-2 signal must first build up the adapter overlaps $\mPhi$ and $\mGamma$ through the rank-$L$ bottleneck, while the up-projection $\bm B$ evolves on the slower readout timescale. This transient delays the growth of the Task 2 overlap, as seen in Fig.~\ref{fig1:StandardVSLoRA}d, and slows early adaptation relative to full fine-tuning.

\section{Stability and Plasticity through the Lens of the Theory}

\subsection{State-Dependent Masking Mitigates Forgetting}
\label{sec:sdgm}

\begin{figure}[t!]
    \includegraphics[width=0.99\linewidth]{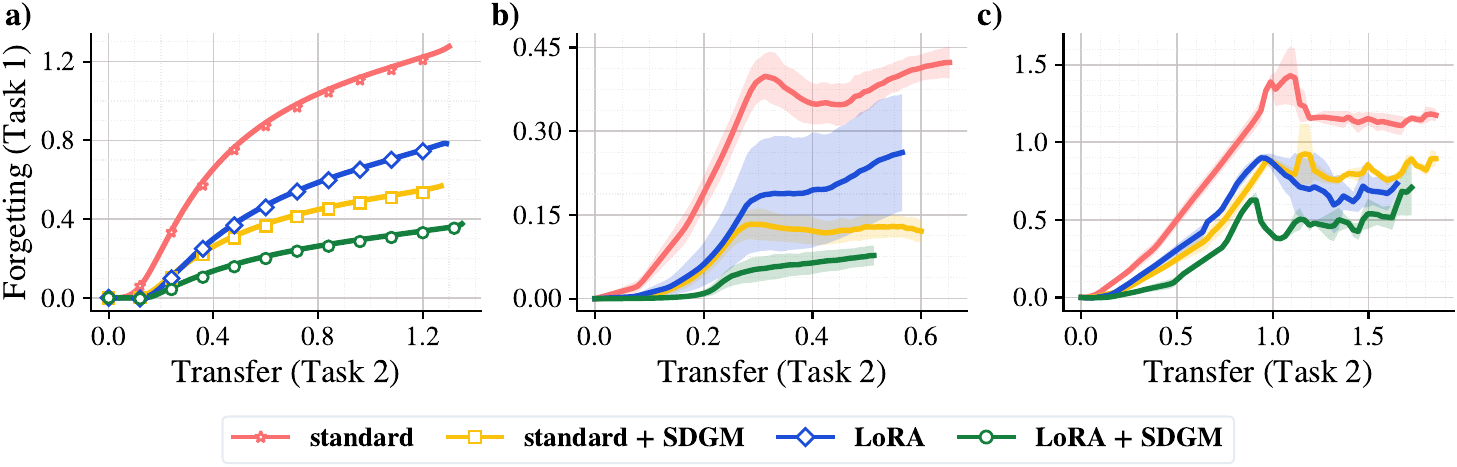}
    \caption{\textbf{State-dependent masking reduces forgetting.} Task 1 forgetting against Task 2 transfer during Task 2 training, with both quantities defined in Eq.~\ref{eq:def_forgetting_transfer}. Since the methods learn at different speeds, forgetting is read at the first time each run reaches a given transfer. \textit{a)} Teacher-student model. Solid lines denote theoretical ODE predictions and markers finite-dimensional simulations. Parameters: $N=10^3$, $K=10$, $M=5$, $L=5$, $\kappa=5$, $c=0.5$, $\alpha=50$. \textit{b)} Sequential MNIST (Appendix~\ref{app:MNIST}), averaged over 10 independent training runs. Curves stop at the largest transfer reached by all runs. Parameters: $N=784$, $K=10$, $L=5$. \textit{c)} Deep benchmark (ResNet-18, Split CIFAR-100), at the best learning rate of each method, with $L=8$. Results over a sequence of six tasks are reported in Appendix~\ref{app:deep_benchmark}. Shaded regions show the standard deviation over 10 runs in b) and three runs in c).}
    \label{fig:protocol}
\end{figure}

The theory suggests that forgetting can be reduced by preventing the LoRA update from acting on feature directions that are strongly used by Task 1. In the multi-head teacher-student model, the magnitude of the
Task 1 readout coefficient $|\vh_i^\dagger|$ provides a simple measure of the importance of hidden unit $i$ for the first task. We therefore rank the hidden units by $|\vh_i^\dagger|$ at the task switch, freeze the $\kappa$ largest, and restrict Task 2 adaptation to the complementary set. We refer to this state-dependent partition as State-Dependent Gradient Masking (SDGM).

Within LoRA, we implement this constraint by fixing the up-projection factor to a sparse mask $\bm\Omega\in\{0,1\}^{K\times L}$ whose nonzero rows correspond only to the plastic hidden units, while optimizing the down-projection $\mA$ and the Task 2 readout $\vh^\ddagger$. The effective representation is therefore

\begin{equation}
\mJ
=
\mJ_s+\frac{\gamma}{\sqrt L}\bm\Omega\mA .
\end{equation}

Because $\bm\Omega$ is constructed from the Task 1 state and then held fixed, the same macroscopic theory applies by setting $\mB=\bm\Omega$ and $d\mB/d\tau=0$, while evolving
$\mPhi,\mXi,\mGamma,\mLambda$ and $\vh^\ddagger$. The explicit mask construction and the corresponding modification of the ODE system are given in Appendix~\ref{sec:SDGM_implementation}.

This construction is closely related to subspace-constrained PEFT methods such as InfLoRA and MiCA \citep{Liang2024,2604.01694}. The distinction is that here the protected subspace is selected directly from the network state reached after Task 1, using the task-specific readout as an importance score, and its effect on the subsequent dynamics can be followed analytically.

Because the methods learn Task 2 at different speeds, comparing their forgetting at a fixed training time would conflate stability with learning speed. We therefore follow both tasks along Task 2 training through

\begin{equation}
\text{Forgetting}(t) = \log \epsilon^\dagger(t) - \log \epsilon^\dagger_s, \qquad
\text{Transfer}(\tau) = \log \epsilon^\ddagger_s - \log \epsilon^\ddagger(t),
\label{eq:def_forgetting_transfer}
\end{equation}

where $\epsilon^*(t)$ is the generalization error on task $*$ at time $t$ and $\epsilon^*_s$ its value at the task switch. Both quantities vanish at the switch, and are positive when the Task 1 error increases and when the Task 2 error decreases, respectively. Plotting forgetting against transfer removes time from the comparison: we read the forgetting of each method at the first time it reaches a given transfer, so that a method cannot appear more stable simply because it learns less.

Figure~\ref{fig:protocol}a shows that, at matched transfer, SDGM (green) substantially reduces Task 1 forgetting relative to vanilla LoRA (blue) while reaching a similar final transfer, with the theory closely matching finite-dimensional simulations. The inverse and random selection controls in Appendix~\ref{app:inv_SDGM}, which protect the least important or randomly chosen Task 1 units instead, produce substantially more forgetting. This shows that the gain depends on \emph{which} directions are protected rather than only on reducing the dimensionality of the trainable subspace. The same qualitative effect persists with ReLU activations as illustrated in Appendix~\ref{app:ReLU_trial}. In the specialized regime presented in Appendix~\ref{app:specialization}, the advantage of vanilla LoRA over full fine-tuning disappears, while SDGM still reduces forgetting at matched Task 2 performance.

Applying the same state-dependent mask to standard fine-tuning (standard + SDGM, yellow) also reduces forgetting, confirming that targeted protection of Task 1-relevant features is beneficial independently of the low-rank parameterization (Fig.~\ref{fig:protocol}a).
 However, in the regime considered here, combining this restriction with LoRA (green) yields the strongest Task 1 retention at comparable Task 2 performance.

To test whether these mechanisms survive beyond the solvable model, we apply the same protocol to two settings increasingly far from the theory. As in the theory, we compare the methods at matched transfer (Eq.~\ref{eq:def_forgetting_transfer}). On sequential MNIST (Fig.~\ref{fig:protocol}b), the two-layer network of the theory is trained online on real data, first to separate digits below and above 5 (Task 1), then even from odd digits (Task 2), as detailed in Appendix~\ref{app:MNIST}. In the deep benchmark (Fig.~\ref{fig:protocol}c), a ResNet-18 is first trained from scratch to classify 50 classes of CIFAR-100 (Task 1), and then learns to classify 10 new classes (Task 2) with each of the four methods. LoRA adapters with rank $L=8$ are placed on every convolution, and SDGM acts on the output channels of each layer, which play the role of the hidden units of the theory. In both settings, the ordering predicted by the theory carries over: full fine-tuning forgets the most, standard + SDGM and LoRA forget less, and LoRA + SDGM forgets the least. Appendix~\ref{app:deep_benchmark} shows the same ordering over a sequence of six tasks, together with measurements of where the update falls and how the Task 1 representation drifts, which connect these results to the mechanism of the theory.

\subsection{Adapter Rank and the Stability--Plasticity Trade-off}
\label{sec:rank}

The LoRA rank $L$ controls the dimensionality of the trainable update and therefore provides a natural handle on the stability-plasticity trade-off. Our dynamical framework quantifies this trade-off by varying $L$ while tracking both transfer to Task 2 and forgetting on Task 1. In Fig.~\ref{fig:forgetting+transfer}, both quantities are taken at the end of the sequential training, i.e.\ Eq.~\ref{eq:def_forgetting_transfer} evaluated at the final time. Figure~\ref{fig:forgetting+transfer}c shows that, for vanilla LoRA (blue), increasing $L$ initially improves Task 2 transfer but also increases Task 1 forgetting. With SDGM (green), transfer likewise improves with rank, while forgetting remains substantially lower because adaptation is restricted away from Task 1-relevant directions. For this comparison, we set $\kappa=K-L$ so that the number of frozen directions decreases with the adapter capacity. At the same time, this choice allows the student to allocate exactly $L$ units for Task 2.

A second feature is that, in the unspecialized regime and at the training horizon considered, transfer gains saturate as $L$ approaches the teacher width $M$. Since Task 2 is generated by $M$ independent feature directions, increasing the rank beyond this scale provides little additional representational benefit on Task 2, while retaining increasingly less information on Task 1, resulting in increased forgetting. The specialized regime gives a sharp version of this statement: LoRA + SDGM escapes the Task 2 plateau only if $\min(L, K - \kappa) \geq M$ as shown in Appendix~\ref{sec:fixed_kappa}.

\subsection{Task similarity and interference}
\label{sec:task_similarity}

\begin{figure}[t!]
    \centering
    \includegraphics[width=0.95\linewidth]{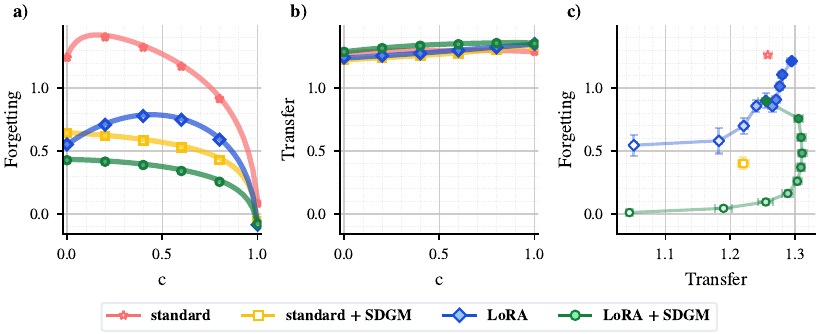}
    \caption{\textbf{Impact of LoRA on  the Stability-Plasticity Trade-off.} \textit{a)} Task 1 Forgetting and \textit{b)} Task 2 Transfer as a function of teacher similarity $c$. Solid lines denote theoretical predictions and markers simulation. \textit{c)} Forgetting versus Transfer as the LoRA rank L varies from light ($L=1$) to dark marker ($L=10$) for $c=0.5$. Vanilla LoRA gains transfer at the cost of increased forgetting, whereas LoRA+SDGM maintains greater stability. In this regime, transfer gains saturate as $L$ approaches the teacher width $M$. For SDGM, $\kappa = K-L$ while the rank sweep at fixed $\kappa$ is reported in Appendix~\ref{sec:fixed_kappa}. We average over 10 different seeds. Lines in panel \textit{c} are shown to guide the eyes. Parameters: $N=10^3$, $K=10$, $M=5$, $\alpha=50$. In panel \textit{a} and \textit{b} we used $L=5$.}
    \label{fig:forgetting+transfer}
\end{figure}

We next vary the teacher similarity $c\in[0,1]$. As shown in Fig.~\ref{fig:forgetting+transfer}a, in the unspecialized regime, forgetting under full fine-tuning (red) and vanilla LoRA (blue) is strongly non-monotonic, with maximal interference at intermediate similarity, consistent with previous continual-learning analyses
\citep{Ramasesh2021,lee2021continual,lee2022,jarvis2025}. When $c$ is small, the tasks occupy nearly orthogonal feature directions and interact weakly; when $c$ approaches one, previously learned features can be reused. At intermediate similarity, however, the tasks overlap enough to induce updates along shared directions while remaining sufficiently
different to distort the Task 1 representation.

SDGM substantially suppresses this intermediate-similarity interference
while preserving comparable Task 2 transfer across the range of $c$ (panel b). This highlights the key geometric limitation of vanilla LoRA: restricting the \emph{rank} of the update does not control its \emph{orientation} relative to previously learned features. By protecting Task 1-relevant directions and redirecting adaptation toward the complementary subspace, SDGM improves the stability-plasticity trade-off especially when the update is already low rank. The deep benchmark shows the same limitation: the share of the LoRA update that falls on the channels most important for Task 1 stays at chance level at every rank, see Fig.~\ref{fig:deep_share}a of Appendix~\ref{app:deep_benchmark}. In the specialized regime shown in Appendix~\ref{app:specialization}, at matched transfer, the forgetting of full fine-tuning and LoRA instead decreases with c, while SDGM keeps it nearly independent of c and is systematically smaller.

\section{Discussion and Conclusions}

We developed a high-dimensional dynamical theory of LoRA in continual learning that follows the complete sequential process from Task 1 feature learning to Task 2 low-rank adaptation. The theory shows that LoRA changes both the geometry and timescale of learning: freezing the pretrained representation reduces interference as long as the Task 1 representation is not specialized, whereas the rank-restricted, zero-initialized adapter must first build up alignment through the low-rank bottleneck, which slows early adaptation. Within this model, low rank alone does not prevent forgetting; the orientation of the adaptation subspace relative to previously learned features is equally important. The specialized regime makes this explicit. There, LoRA forgets as much as full fine-tuning, and only state-dependent masking protects Task 1. This perspective explains why state-dependent masking reduces forgetting and clarifies how rank and task similarity shape the stability-plasticity trade-off.

Our analysis is deliberately restricted to a solvable two-layer, two-task online-learning model, so its quantitative predictions should
not be transferred directly to large deep networks. Its purpose is instead to isolate mechanisms that are difficult to disentangle
empirically. The qualitative agreement on sequential MNIST and Split CIFAR-100, using a shallow network and a deep convolutional network (ResNet-18), respectively, suggests that these mechanisms extend beyond the analytically tractable setting and motivates further study of dynamically constrained adaptation in deeper networks and longer task sequences.

\subsubsection*{Acknowledgments}
We thank Sebastian Goldt, Stefano Sarao Mannelli and Francesco Camilli for insightful discussions on this work. The work of TM was supported by the European Union – NextGenerationEU under the National Recovery and Resilience Plan (PNRR), Mission 4, Component 2, Investment 3.3, “Introduction of innovative PhD programmes responding to the innovation needs of enterprises and promoting the recruitment of researchers by enterprises” (D.M. 630/2024), CUP J33C24001630009, and by Syndiag S.r.L. This work was conducted in the spirit of the Slow Science Manifesto \href{https://www.slow-science.com/}{slow-science.com}, advocating for collaborative and sustainable research.

\bibliography{iclr2027_conference}
\bibliographystyle{iclr2027_conference}

\appendix

\section{Integrals computation}
\label{app:integrals_computation}
To derive the closed-form equations for the various quantities we will determine, we make use of several quantities involving Gaussian integrals. For completeness, we report their expressions in this Appendix, following \citet{Saad1995}.We denote by
\begin{equation}
\langle f(\vx) \rangle_{\vx}
\equiv
\int d\vx\, P(\vx) f(\vx),
\end{equation}
the expectation of a function $f$ with respect to its joint Gaussian distribution with zero mean and covariance matrix $\mC$. All the computations that follow are specific to $g(x)=\erf(x/\sqrt{2})$.

For the two-dimensional case, we define
\begin{equation}
I_2 = \left\langle g(x)g(y)\right\rangle_{(x,y)}.
\end{equation}
This Gaussian integral admits the closed-form expression
\begin{equation}
\label{eq:I2_integral}
I_2(a,b) = \frac{2}{\pi} \arcsin\left( 
    \frac{\mC_{ab}} {\sqrt{\mC_{aa}+1}\sqrt{\mC_{bb}+1}}
\right).
\end{equation}
Likewise, the computation of a three-dimensional gaussian average 
\begin{equation} I_3 = \langle g'(x)\, y \, g(z) \rangle_{(x,y,z)} \end{equation}
can be computed to get
\begin{equation}
\label{eq:I3_integral}
I_3(a,b,c) = \frac{2}{\pi}\frac{1}{\sqrt{\Lambda_3(a,c)}} \frac{ \mC_{bc}(\mC_{aa}+1)-\mC_{ab} \mC_{ac} }{ \mC_{aa} + 1 }.
\end{equation}

Finally, the four-dimensional gaussian integral
\begin{equation}
I_4 = \langle g'(x)\, g'(y)\, g(z)\, g(w)\rangle_{(x,y,z,w)} 
\end{equation}
results in
\begin{equation}
\label{eq:I4_integral}
I_4(a,b,c,d) = \frac{4}{\pi^2\sqrt{\Lambda_4(a,b)}}\arcsin{\left( \frac{\Lambda_0(a,b,c,d)}{\sqrt{\Lambda_1(a,b,c)}\sqrt{\Lambda_2(a,b,d)}} \right)}.
\end{equation}
Where we defined
\begin{align*}
    \Lambda_4(a,b) & := (\mC_{aa}+1)(\mC_{bb}+1)-\mC_{ab}^2, \\
    \Lambda_0(a,b,c,d) & := \Lambda_4 \mC_{cd} - \mC_{ac} \mC_{ad}( \mC_{bb}+1) - \mC_{bc} \mC_{bd}(\mC_{aa}+1) + \mC_{ab}(\mC_{ad} \mC_{bc} + \mC_{bd}\mC_{ac}), \\
    \Lambda_1(a,b,c) & := \Lambda_4(\mC_{cc}+1) - \mC_{ac}^2(\mC_{bb}+1) - \mC_{bc}^2(\mC_{aa}+1)+2 \mC_{ab} \mC_{ac} \mC_{bc}, \\
    \Lambda_2(a,b,d) & := \Lambda_4(\mC_{dd}+1) - \mC_{ad}^2( \mC_{bb}+1) - \mC_{bd}^2(\mC_{aa}+1)+2 \mC_{ab} \mC_{ad} \mC_{bd},\\
    \Lambda_3(a,c) & := (\mC_{aa}+1)(\mC_{cc}+1)-\mC_{ac}^2.
\end{align*}

\section{Standard sequential training}
\label{app:theory_task1}
 The results presented for the training on Task 1 are identical to those presented in \citet{lee2021continual,lee2022} and also apply to the first phase of this LoRA-based framework. We choose to state them in this Appendix for completeness. We also report the theoretical ODEs needed to reproduce the standard approach when training on Task 2, which serve as a comparison with the theoretical results found in this work.
\subsubsection*{List of order parameters}

We start by defining the preactivation fields of the $m^{th}$ teacher $\dagger$ unit, $p^{th}$ teacher $\ddagger$ unit and $k^{th}$ student unit respectively as
\begin{equation}
    \rho_m = \frac{\mW^\dagger_m \vxi}{\sqrt{N}}, \quad     \nu_p = \frac{\mW^\ddagger_p \vxi}{\sqrt{N}}, \quad     x_k = \frac{\mJ_k \vxi}{\sqrt{N}}. 
\end{equation}

The set of time-dependent order parameters we recover during the first part of training are then 
\begin{align}
& \textrm{Student - Student Overlap, } \mQ_{kl} := \langle x_k x_l\rangle = \frac{1}{N}\mJ_k \mJ^T_l,\\
& \textrm{Student - Teacher}^\dagger \textrm{ Overlap, } \mR_{km} := \langle x_k \rho_m \rangle = \frac{1}{N} \mJ_k (\mW^\dagger_m)^T,\\
& \textrm{Student - Teacher}^\ddagger \textrm{Overlap, } \mU_{kp} := \langle x_k \nu_p \rangle = \frac{1}{N} \mJ_k (\mW^\ddagger_p)^T.
\end{align}
While the static order parameters, totally defined by the sampling procedure for the teachers given in~\ref{sec:LoRA_problem_setting} are given by
\begin{align}
    & \textrm{Teacher}^\dagger \textrm{ - Teacher}^\dagger \textrm{ Overlap, } \mT_{nm} := \langle \rho_n \rho_m\rangle = \frac{1}{N} \mW^\dagger_n \left( \mW^\dagger \right)^T_m, \\
    & \textrm{Teacher}^\ddagger \textrm{ - Teacher}^\ddagger \textrm{ Overlap, } \mS_{pq} := \langle \nu_p \nu_q \rangle = \frac{1}{N} \mW^\ddagger_p \left( \mW^\ddagger \right)^T_q, \\
    & \textrm{Teacher}^\dagger \textrm{ - Teacher}^\ddagger \textrm{ Overlap, } \mV_{mp} := \langle \rho_m \nu_p\rangle = \frac{1}{N} \mW^\dagger_m \left( \mW^\ddagger \right)^T_p.
\end{align}

\subsubsection*{Generalization errors}
Having defined the order parameters and preactivation fields, we are now ready to compute the generalization error on both tasks during the first part of training. 

Recall that computing an average over the distribution of the input $\vxi\sim\mathcal{N}(0,\mathbf{I}_N)$ is not needed when the functions involved depend only on the preactivations. This means we want to focus on the joint distribution of such preactivations fields
\begin{equation}
\label{gaussianvector_task1}
(x_1, \ldots ,x_K, \rho_1 , \ldots ,\rho_M,  \nu_1 , \ldots , \nu_M ).
\end{equation}
In the limit $N \to \infty$, the preactivations fields are jointly gaussian and we need only to focus on the time-dependent joint covariance matrix of those preactivations, written as
\begin{equation}
    \label{eq:covarianceMatrixClassical}
    \tilde{\mC} = \begin{bmatrix}
        \mQ & \mR & \mU\\
        \mR^T & \mT & \mV\\
        \mU^T & \mV^T & \mS
    \end{bmatrix}.
\end{equation}

We first express the generalization errors on both tasks in term of the preactivations, giving
\begin{align}
    &\epsilon^\dagger = \frac{1}{2} \Biggl< \sum_{k,l=1}^K h_k^\dagger h_l^\dagger g(x_k) g(x_l) + \sum_{m,n=1}^M \vv_m^\dagger \vv_n^\dagger g(\rho_m) g(\rho_n) - 2 \sum_{k=1}^K \sum_{m=1}^M \vh_k^\dagger \vv_m^\dagger g(x_k) g(\rho_m)\Biggr>, \\
    & \epsilon^\ddagger = \frac{1}{2} \Biggl< \sum_{k,l=1}^K h_k^\ddagger h_l^\ddagger g(x_k) g(x_l) + \sum_{p,q=1}^M \vv_p^\ddagger \vv_q^\ddagger g(\nu_p) g(\nu_q) - 2 \sum_{k=1}^K \sum_{p=1}^M \vh_k^\ddagger \vv_p^\ddagger g(x_k) g(\nu_p)\Biggr>.
\end{align}
Using the Gaussian integrals, we obtain a closed form solution for the errors
\begin{align}
    &\epsilon^\dagger = \frac{1}{2}  \sum_{k,l=1}^K \vh_k^\dagger \vh_l^\dagger I_2(k,l) + \frac{1}{2}\sum_{m,n=1}^M \vv_m^\dagger \vv_n^\dagger I_2(K+m,K+n) -  \sum_{k=1}^K \sum_{m=1}^M \vh_k^\dagger \vv_m^\dagger I_2(k,K+m), \\
    & \epsilon^\ddagger = \frac{1}{2} \sum_{k,l=1}^K \vh_k^\ddagger \vh_l^\ddagger I_2(k,l) + \frac{1}{2}\sum_{p,q=1}^M \vv_p^\ddagger \vv_q^\ddagger I_2(K+M+p,K+M+q) -  \sum_{k=1}^K \sum_{p=1}^M \vh_k^\ddagger \vv_p^\ddagger I_2(k,K+M+p).
\end{align}
where the variables of the various gaussian integrals represent the associated entries of the covariance matrix (\ref{eq:covarianceMatrixClassical}). 
\subsubsection*{Gradient updates}
In this section, we report the weight updates while training on the two tasks in the classical setting of~\citet{lee2021continual}.
In this case, the output of the student network on task $*\in\{\dagger,\ddagger\}$ is given by
\begin{equation}
\label{outputTask1}
\phi(\vxi^\mu; \mJ, \vh^*) = \sum_{k=1}^K { {\vh^*_k} g \left( \frac{\mJ_k \vxi^{\mu}}{\sqrt{N}} \right) }.
\end{equation}

The loss on input $\vxi^{\mu}$ explicitly reads
\begin{equation}
\label{lossSTANDARD}
\ell( \vxi^\mu ;\mJ ,\vh^*, \mW^*,  \vv^*  ) = \frac{1}{2} \Biggl( \sum_{m=1}^{M}{\vv^*_m\,g \biggl( \frac{\mW^*_m\vxi^{\mu}}{\sqrt{N}}\biggr)} - \sum_{k=1}^{K}{{\vh^*_k}\,g \biggl( \frac{\mJ_k  \vxi^{\mu}}{\sqrt{N}} \biggr)} \Biggr)^2.
\end{equation}
We define the prediction error on the $\mu$-th example for task $*$ as
\begin{equation}
\Delta^{*,\mu}
=
\sum_{k=1}^{K}
\bigl(\vh_k^*\bigr)^\mu g(x_k^\mu)
-
\sum_{m=1}^{M}
\vv_m^* g^*_m, \quad g^\dagger_m:=g(\rho_m) ; \, \, g^\ddagger_m:= g(\nu_m).
\end{equation}
This allows us to compute the gradient with respect to the $i$-th row of $\mJ$
\begin{equation}
    \label{eq:gradient_J}
    \nabla_{\mJ_i}\ell( \vxi^\mu ;\mJ ,\vh^*, \mW^*,  \vv^*  ) = \Delta^{*,\mu} \bigl(\vh^*_i \bigr)^{\mu}g'(x_i^{\mu})\frac{(\vxi^{\mu})^T}{\sqrt{N}}
\end{equation}
from which the gradient update for $\mJ_i$ reads
\begin{equation}
    \label{eq:updateJ}
    \mJ_{i}^{\mu+1} = \mJ_{i}^{\mu}- \frac{\eta_{\mJ}}{\sqrt{N}} \Delta^{*,\mu}\bigl({\vh^*_i}\bigr)^{\mu}g'(x_i^{\mu})(\vxi^{\mu})^T.
\end{equation}
In the same fashion, the gradient update for $\vh^*$ is given by
\begin{equation}
    \label{eq:updateH}
    \bigl({\vh_{i}^*}\bigr)^{\mu+1} = \bigl({\vh_{i}^*} \bigr)^{\mu}- \frac{\eta_{\vh}}{N} \Delta^{*,\mu}g(x_i^{\mu}).
\end{equation}
\subsubsection*{Differential equations}
In the following, we will make use of the following notation:
\begin{align}
    \label{eq:Delays}
    \begin{split}
        &\mathrm{delay}^\dagger = 0, \\
        &\mathrm{delay}^\ddagger = M.
\end{split}
\end{align}

All integrals computed in this section are performed on the covariance matrix of standard training~(\ref{eq:covarianceMatrixClassical}). The delays~(\ref{eq:Delays}) are needed because the relevant preactivation fields of the teacher in the covariance matrix depend on the active task $*\in\{\dagger,\ddagger\}$ we are considering when solving the ODEs.\\
\paragraph{$\mQ$:} From the gradient update~(\ref{eq:updateJ}), multiplying by $(\mJ_k^{\mu+1})^T$ on the right and using the corresponding expressions for the preactivations:

\begin{align*}
    \begin{split}
    \mJ_i^{\mu+1}(\mJ_k^{\mu+1})^T  = &\mJ_{i}^{\mu}  (\mJ_{k}^{\mu})^T - \eta_\mJ\,\Delta^{*,\mu}\, \bigl(\vh^*_{i}\bigr)^{\mu} \, g'(x_i^{\mu})\,x_k^{\mu} - \eta_\mJ\,\Delta^{*,\mu}\, \bigl(\vh^*_{k}\bigr)^{\mu} \, g'(x_k^{\mu})\,x_i^{\mu} \\
    & + \eta_\mJ^2\bigl(\Delta^{*,\mu}\bigr)^2\, \bigl(\vh^*_{i}\bigr)^{\mu} \, g'(x_i^{\mu})\,\bigl(\vh^*_{k}\bigr)^{\mu} \, g'(x_k^{\mu})\,\frac{||\vxi^{\mu}||^2}{N}.
    \end{split}
\end{align*}
Rearranging terms and substituting the order parameters, we obtain
\begin{align*}
    \begin{split}
    \frac{ \mQ_{ik}^{\mu+1} - \mQ_{ik}^{\mu}}{1/N} & = - \eta_\mJ\,\Delta^{*,\mu}\, \bigl(\vh^*_{i}\bigr)^{\mu} \, g'(x_i^{\mu})\,x_k^{\mu} - \eta_\mJ\,\Delta^{*,\mu}\, \bigl(\vh^*_{k}\bigr)^{\mu} \, g'(x_k^{\mu})\,x_i^{\mu} \\
    & + \eta_\mJ^2(\Delta^{*,\mu})^2\, \bigl(\vh^*_{i}\bigr)^{\mu} \, g'(x_i^{\mu})\,\bigl(\vh^*_{k}\bigr)^{\mu} \, g'(x_k^{\mu})\,\frac{||\vxi^{\mu}||^2}{N}.
    \end{split}
\end{align*}
Defining $\tau:=\mu/N$ and taking the thermodynamic limit
$N\to\infty$, the discrete difference equation converges
to the continuous-time differential equation:
\begin{equation*}
    \frac{\mathrm{d}\mQ_{ik}}{\mathrm{d}\tau} = -\eta_\mJ\,\vh^*_i\,\langle g'(x_i)\,x_k\,\Delta^* \rangle - \eta_\mJ\,\vh^*_k\,\langle g'(x_k)\,x_i\,\Delta^* \rangle + \eta_\mJ^2\,\vh^*_i\,\vh^*_k\,\langle g'(x_i)g'(x_k)(\Delta^*)^2 \rangle
\end{equation*}
that we can rewrite in terms of the integrals found in Appendix~\ref{app:integrals_computation} after explicitly substituting $\Delta^*$ and $(\Delta^*)^2$:
\begin{align}
    \label{eq:ODE_forQ_integral}
    \begin{split}
    \frac{\mathrm{d}\mQ_{ik}}{\mathrm{d}\tau} = \, &\eta_\mJ\,\vh^*_i\left[ \sum_{m=1}^{M}\vv^*_m\,I_3(i,k,K+\mathrm{delay}^*+m) - \sum_{j=1}^{K}\vh^*_j\,I_3(i,k,j) \right] \\
    +\,& \,\eta_\mJ\,\vh^*_k\left[ \sum_{m=1}^{M}\vv^*_m\,I_3(k,i,K+\mathrm{delay}^*+m) - \sum_{j=1}^{K}\vh^*_j\,I_3(k,i,j) \right] \\
    +\,&\eta_\mJ^2\,\vh^*_i\,\vh^*_k \left[ \sum_{j,l=1}^{K} \vh^*_j \, \vh^*_l \,I_4(i,k,j,l) \right. \\
    +\,&\sum_{m,n=1}^{M} \vv^*_m \, \vv^*_n \,I_4(i,k,K+\mathrm{delay}^*+m,K+\mathrm{delay}^*+n) \\
    -\,&\left.2 \sum_{j=1}^{K}\sum_{m=1}^{M} \vh^*_j \, \vv^*_m \, \,I_4(i,k,j,K+\mathrm{delay}^*+m)\right].
    \end{split}
\end{align}
\paragraph{$\mR$:} From the gradient update~\ref{eq:updateJ}, multiplying by ${\mW^\dagger_n}^T$ on the right and using the corresponding expressions for the relevant preactivations:
\begin{equation*}
    \mJ_i^{\mu+1}{\mW^\dagger_n}^T = \mJ_{i}^{\mu}  {\mW^\dagger_n}^T - \eta_\mJ\,{\Delta^*}^{\mu}\, {\vh^*_{i}}^{\mu} \, g'(x_i^{\mu})\,\rho_n^{\mu}
\end{equation*}
Rearranging terms and substituting the order parameter, we obtain:
\begin{equation*}
    \frac{ \mR_{in}^{\mu+1} - \mR_{in}^{\mu}}{1/N} = - \eta_\mJ \, \Delta^{*,\mu} \, \bigl(\vh^*_{i}\bigr)^{\mu} \, g'(x_i^{\mu}) \, \rho_n^{\mu}.
\end{equation*}
Performing the thermodynamic limit $N\to\infty$ we obtain the differential equation:
\begin{equation}
    \label{eq:ODE_forR_integral}
    \frac{\mathrm{d}\mR_{in}}{\mathrm{d}\tau} = \eta_\mJ\,{\vh^*_{i}}\,\left[ \sum_{m=1}^{M}{{\vv^*_{m}}\,I_3(i,K+n,K+\mathrm{delay}^*+m)} - \sum_{j=1}^{K}{{\vh^*_{j}}\,I_3(i,K+n,j) }\right],
\end{equation}
\paragraph{$\vh^*$:} From the gradient update~\ref{eq:updateH} we can write
\begin{equation*}
    \frac{{\vh^*_{i}}^{\mu+1} - {\vh^*_{i}}^{\mu}}{1/N} = - \eta_{\vh}\Delta^{*,\mu}g(x_i^{\mu})
\end{equation*}
and take the thermodynamic limit to obtain
\begin{equation}
    \label{eq:ODE_forh_integral}
    \frac{\mathrm{d}\vh^*_i}{\mathrm{d}\tau} = \eta_\vh\,\left[ \sum_{m=1}^{M}{{\vv^*_{m}}\,I_2(K+\mathrm{delay}^*+m,i)} - \sum_{j=1}^{K}{{\vh^*_{j}}\,I_2(j,i)}\right].
\end{equation}
\paragraph{$\mU$:} From the gradient update~\ref{eq:updateJ}, multiplying by ${\mW^\ddagger_p}^T$ on the right and using the corresponding expressions for the relevant preactivations:
\begin{equation*}
    \mJ_i^{\mu+1}{\mW^\ddagger_p}^T = \mJ_{i}^{\mu}  {\mW^\ddagger_p}^T - \eta_\mJ\,\Delta^{*,\mu}\, \bigl(\vh^*_{i}\bigr)^{\mu} \, g'(x_i^{\mu})\,\nu_p^{\mu}
\end{equation*}
Rearranging terms and substituting the order parameter, we obtain:
\begin{equation*}
    \frac{ \mU_{ip}^{\mu+1} - \mU_{ip}^{\mu}}{1/N} = - \eta_\mJ \, \Delta^{*,\mu} \, \bigl(\vh^*_{i}\bigr)^{\mu} \, g'(x_i^{\mu}) \, \nu_p^{\mu}.
\end{equation*}
Performing the thermodynamic limit $N\to\infty$ we obtain the differential equation:
\begin{equation}
    \label{eq:ODE_forU_integral}
    \frac{\mathrm{d}\mU_{ip}}{\mathrm{d}\tau} = \eta_\mJ\,\vh^*_i\,\left[ \sum_{m=1}^{M}{\vv^*_m\,I_3(i,K+M+p,K+\mathrm{delay}^*+m)} - \sum_{j=1}^{K}{\vh^*_j\,I_3(i,K+M+p,j) }\right].
\end{equation}
\section{Low-rank Adaptation on Task 2}
This section presents the new differential equations arising from Low-Rank Adaptation in the online learning paradigm. The equations given in this appendix must be used when training on Task 2 only. To retrieve $\mQ_s,\mR_s,\mU_s,\vh^\dagger$, the ODEs given in Appendix~\ref{app:theory_task1} must first be integrated (by setting $*=\dagger)$. In the following, we use the shortcut $\beta:=\gamma/\sqrt{L}$.
\label{app:theory_Task2}
\subsubsection*{List of order parameters}
By defining the preactivation fields of the $m^{th}$ teacher $\dagger$ unit, $p^{th}$ teacher $\ddagger$ unit, $k^{th}$ student unit, and $i^{th}$ direction of the LoRA update of the student unit respectively as
\begin{equation}
    \rho_m = \frac{\mW^\dagger_m \vxi}{\sqrt{N}}, \quad     \nu_p = \frac{\mW^\ddagger_p \vxi}{\sqrt{N}}, \quad     x_k = \frac{(\mJ_s)_k \vxi}{\sqrt{N}}, \quad z_i = \frac{\mA_i \vxi}{\sqrt{N}}, \quad     \tilde{x}_k = x_k + \beta \sum_{l=1}^L\mB_{kl} z_l.
\end{equation}
Note that after the task switch, $x_k$ is a constant depending on the frozen weight $(\mJ_s)_k$ only.\\
Hence, the full set of time-dependent order parameters recovered by the theory in the second part of training are:
\begin{align*}
&\text{Down-projection--Down-projection Overlap, }: \mPhi_{ij} := \langle z_i z_j \rangle
= \frac{1}{N}\mA_i \mA_j^T \\
&\text{Teacher}^\dagger\text{--Down-projection Overlap, }: \mLambda_{mi} := \langle \rho_m z_i \rangle
= \frac{1}{N}\mW^\dagger_m\mA_i^T \\
&\text{Teacher}^\ddagger\text{--Down-projection Overlap, }: \mGamma_{pi} := \langle \nu_p z_i \rangle
= \frac{1}{N}\mW^\ddagger_p\mA_i^T \\
&\text{Student--Down-projection Overlap, }:
\mXi_{ki} := \langle x_k z_i \rangle
= \frac{1}{N}(\mJ_s)_k\mA_i^T \\
&\text{Up-projection Matrix, } \mB.
\end{align*}
while the other order parameters 
\begin{align*}
    & \mQ_{kl} := \langle x_k x_l\rangle, \\
    & \mR_{km} := \langle x_k \rho_m \rangle,\\
    & \mU_{kp} := \langle x_k\nu_p \rangle, \\
    & \mT_{mn} := \langle \rho_m \rho_n \rangle,\\
    & \mS_{pq} := \langle \nu_p \nu_q \rangle, \\
    & \mV_{mp} := \langle \rho_m \nu_p \rangle
\end{align*}
are frozen in this specific part of training, their evolution being tracked in the first part of training using the closed form formulae given in \citet{lee2021continual} and mentioned in Appendix~\ref{app:theory_task1}.

\subsubsection*{Generalization errors}
\label{orderparametersLORA}
We start by computing the generalization error on both tasks during the second part of training. 

Recall that computing an average over the distribution of the input $\vxi\sim\mathcal{N}(0,\mathbf{I}_N)$ is not needed when the functions involved depend only on the preactivations. This means we want to focus on the joint distribution of such preactivations. Although $\tilde{x}_i$ is just a linear transformation of $(x_i, z_1, \cdots, z_L)$, it is recommended to consider the higher dimensional gaussian distribution of the random vector
\begin{equation}
\label{newgaussianvector}
(x_1, \ldots ,x_K, \tilde{x}_1 , \ldots , \tilde{x}_K , \nu_1 , \ldots , \nu_M , z_1 , \ldots , z_L, \rho_1 , \ldots ,\rho_M).
\end{equation}
Starting from the definition
\begin{align*}
    \label{oldorderparameters}
    &\langle x_k x_l\rangle = \mQ_{kl}, \\
    &\langle x_k\nu_p \rangle =  \mU_{kp}, \\
    &\langle x_k \rho_m \rangle = \mR_{km} \\
    &\langle \nu_p \nu_q \rangle = \mS_{pq}, \\
    &\langle \rho_m \rho_n \rangle = \mT_{mn} \\
    &\langle \rho_m \nu_p \rangle = \mV_{mp} \\
    &\langle z_i z_j \rangle = \mPhi_{ij}, \\
    &\langle \nu_p z_i \rangle = \mGamma_{pi}, \\
    &\langle \rho_m z_i \rangle = \mLambda_{mi} \\
    &\langle x_k z_i \rangle = \mXi_{ki},
\end{align*}

and computing the remaining interactions
\begin{align*}
    \langle \tilde{x}_k z_i \rangle &= \frac{(\mJ_k+\beta\,\mB_k \mA)\mA_i^T}{N} = \frac{\mJ_k \mA_i^T}{N} + \beta\,\mB_k \frac{\mA\mA_i^T}{N} = \mXi_{ki} + \beta\,\mB_k\,\mPhi_i^T, \\
    \langle \tilde{x}_k \nu_p \rangle &= \frac{(\mJ_k+\beta \, \mB_k \mA) {\mW^{\ddagger}_p}^T}{N} = \frac{\mJ_k {\mW^{\ddagger}_p}^T }{N} + \beta \,\mB_k\frac{\mA{\mW^{\ddagger}_p}^T}{N}= \mU_{kp} + \beta \mB_k \mGamma_p^T,\\
    \langle \tilde{x}_k \rho_m \rangle &= \frac{(\mJ_k+\beta \, \mB_k \mA) {\mW^{\dagger}_m}^T}{N} = \frac{\mJ_k {\mW^{\dagger}_m}^T }{N} + \beta \,\mB_k\frac{\mA{\mW^{\dagger}_m}^T}{N}= \mR_{km} + \beta \mB_k \mLambda_m^T, \\
    \langle x_k\tilde{x}_l \rangle & = \frac{\mJ_k(\mJ_l+\beta \mB_l\mA)^T}{N}=\frac{\mJ_k\mJ_l^T}{N}+\beta\frac{\mJ_k\mA^T\mB_l^T}{N} = \mQ_{kl} + \beta\, \mXi_{k:}\,\mB_l^T, \\
    \begin{split}
        \langle \tilde{x}_k \tilde{x}_l \rangle & = \frac{\tilde{\mJ}_k\tilde{\mJ}_l^T}{N} = \frac{1}{N}\left( \mJ_k + \beta\,\mB_k \mA \right)\left( \mJ_l^T + \beta\,\mA^T\mB_l^T\right) \\
        & = \frac{\mJ_k \mJ_l^T}{N} + \beta\,\frac{\mJ_k \mA^T}{N} \mB_l^T + \beta\,\mB_k\,\frac{\mA\mJ_l^T}{N}+\beta^2\,\mB_k\,\frac{\mA\mA^T}{N}\mB_l^T \\
        & = \mQ_{kl} + \beta\, \mXi_k\mB_l^T + \beta\,\mB_k\mXi_l^T+\beta^2\,\mB_k\mPhi \mB_l^T,
    \end{split}
\end{align*}
allows us to determine the covariance matrix of the ($2K+2M+L$)-dimensional gaussian vector (\ref{newgaussianvector})
\begin{equation}
    \label{covarianceLORA}
    \mC = \begin{bmatrix}
        \mQ & \mQ + \beta\,\mXi\,\mB^T & \mU & \mXi & \mR\\
        \mQ^T + \beta\,\mB\mXi^T & \mQ + \beta(\mXi \mB^T + \mB \mXi ^T) + \beta^2 \mB\mPhi \mB^T &  \mU+\beta\,\mB\mGamma^T & \mXi + \beta\,\mB\mPhi^T & \mR + \beta \mB \mLambda^T \\
        \mU^T & \mU^T + \beta\,\mGamma \mB^T & \mS & \mGamma & \mV^T \\
        \mXi^T & \mXi^T+ \beta\,\mPhi \mB^T & \mGamma^T & \mPhi & \mLambda^T \\
        \mR^T & \mR^T + \beta \mLambda \mB^T & \mV & \mLambda & \mT\\
    \end{bmatrix}.
\end{equation}
We can write the generalization error on the second task in terms of the preactivations:
\begin{equation*} 
\label{lossLORAorders}
\begin{aligned}
\epsilon^\ddagger \left( \vh^\ddagger, \mJ, \vv^\ddagger, \mW^\ddagger \right) &= \frac{1}{2}\Biggl\langle \left( \sum_{p=1}^{M}{\vv_p^\ddagger \,g\biggl(\frac{\mW^\ddagger_p\xi}{\sqrt{N}}\biggr)} - \sum_{k=1}^{K}{\vh_k^\ddagger\,g\left( \frac{\mJ_k  \xi+\beta\,\mB_k \mA\xi}{\sqrt{N}} \right)} \right)^2\Biggr\rangle \\
&= \frac{1}{2}\Biggl\langle \left( \sum_{p=1}^{M}{\vv^\ddagger_p\,g\left(\nu_p\right)} - \sum_{k=1}^{K}{\vh^\ddagger_k\,g\left(\tilde{x}_k\right)} \right)^2 \Biggr\rangle \\
&= \frac{1}{2}\sum_{p,q=1}^M{ \vv^\ddagger_p \vv^\ddagger_q \bigl\langle g(\nu_p)g(\nu_q) \bigr\rangle } \\
& \quad- \sum_{p=1}^M\sum_{k=1}^{K}{ \vv^\ddagger_p \vh^\ddagger_k \bigl\langle g(\nu_p)g(\tilde{x}_k) \bigr\rangle} \\
& \quad + \frac{1}{2} \sum_{k,l=1}^{K}{ \vh^\ddagger_k \vh^\ddagger_l \bigl\langle g(\tilde{x}_k)g(\tilde{x}_l) \bigr\rangle} \\
\end{aligned}
\end{equation*}
Expressing this quantity as a function of the gaussian integrals given in Appendix ~\ref{app:integrals_computation} allows us to cancel the explicit dependency on the first-layers and to close the equation
\begin{equation}
    \begin{aligned}
        \label{generrLORAint}
        \epsilon^\ddagger \left( \vh^\ddagger, \vv^\ddagger \right) &= \frac{1}{2}\sum_{p,q=1}^M{ \vv^\ddagger_p \vv^\ddagger_q I_2(2K+p,2K+q)} \\
        & \quad- \sum_{p=1}^M\sum_{k=1}^{K}{ \vv^\ddagger_p \vh^\ddagger_k I_2(2K+p,K+k)} \\
        & \quad+\frac{1}{2} \sum_{k,l=1}^{K}{ \vh^\ddagger_k \vh^\ddagger_l I_2(K+k,K+l)}.
    \end{aligned}
\end{equation}
In the same fashion, the generalization error on the first task results in
\begin{equation}
    \begin{aligned}
        \label{generrfirsttask}
        \epsilon^\dagger \left( \vh^\dagger, \vv^\dagger \right) &= \frac{1}{2}\sum_{p,q=1}^M{ \vv^\dagger_p \vv^\dagger_q I_2(2K+M+L+p,2K+M+L+q)} \\
        & \quad - \sum_{p=1}^M\sum_{k=1}^{K}{ \vv^\dagger_p \vh^\dagger_k I_2(2K+M+L+p, K+ k)} \\
        & \quad +\frac{1}{2} \sum_{k,l=1}^{K}{ \vh^\dagger_k \vh^\dagger_l I_2(K+k,K+l)}.
    \end{aligned}
\end{equation}

\subsubsection*{Gradient Updates}
We start by deriving the gradient updates of the LoRA adapter.

Let $\vxi^{\mu}$ be the input vector of the network and $\beta:=\gamma/\sqrt{L}$ the LoRA scaling factor. The output of the student network on the second task is given by
\begin{equation}
\label{outputLORA}
\phi(\vxi^\mu; \mJ_s, \mB, \mA , \vh^\ddagger) = \sum_{k=1}^K { {\vh^\ddagger_k}^{\mu} g \left( \frac{(\mJ_s)_k \vxi^{\mu}+\beta \mB_k^{\mu}\mA^{\mu}\vxi^{\mu}}{\sqrt{N}} \right) }.
\end{equation}
Where $(\mJ_s)_k$ and $\mB_k$ are the $k$-th rows of $\mJ_s$ and $\mB$, respectively. We also define $\tilde{\mJ}_k = (\mJ_s)_k + \beta \mB_k\mathbf{\mA}$. The loss on input $\vxi^{\mu}$ can be rewritten as
\begin{equation}
\label{lossLORA}
\ell( \vxi^\mu ;\vh^\ddagger, \mJ_s , \mB , \mA , \vv , \mW^\ddagger ) = \frac{1}{2} \Biggl( \sum_{m=1}^{M}{\vv^\ddagger_m\,g \biggl( \frac{\mW^\ddagger_m\xi^{\mu}}{\sqrt{N}}\biggr)} - \sum_{k=1}^{K}{{\vh^\ddagger_k}^{\mu}\,g \biggl( \frac{\mJ_k  \vxi^{\mu}+\beta \mB_k^{\mu}\mA^{\mu}\vxi^{\mu}}{\sqrt{N}} \biggr)} \Biggr)^2. \end{equation}

Using the preactivations, we can compute the partial derivative of \ref{lossLORA} with respect to $\mB_{ki}$:
\begin{equation}
    \label{partialDjs}
    \begin{aligned}\partial_{\mB_{ki}}
    \ell( \vxi^\mu ;\vh^\ddagger,\mJ_s , \mB , \mA , \vv^\ddagger , \mW^\ddagger ) & = (\Delta^\ddagger)^{\mu}\partial_{\mB_{ki}}\left( \sum_{l=1}^{K}{{\vh^\ddagger_l}^{\mu}\,g\left(x_l^{\mu} + \beta\sum_{j=1}^L{\mB_{lj}^{\mu}z_j^{\mu}}\right) } \right) \\
    & = \beta (\Delta^\ddagger)^{\mu} {\vh^\ddagger_k}^{\mu} g'\left(\tilde{x}_k^{\mu} \right)z_i^{\mu}.
    \end{aligned}
\end{equation}
The gradient update for $\mB$ is therefore
\begin{equation}
    \label{updateD}
    \mB_{ki}^{\mu+1} = \mB_{ki}^{\mu}- \beta \frac{\eta_{\mB}}{N} (\Delta^\ddagger)^{\mu} {\vh^\ddagger_k}^{\mu} g'\left(\tilde{x}_k^{\mu} \right)z_i^{\mu}
\end{equation}
where an extra prefactor $1/N$ has been added in the gradient update rule of $\mB$ in the same fashion as the update rule for the readout weights (\ref{eq:updateH}, \ref{updateH_LORA}), such that the scaling in $1/N$ allows for a well-defined, non-trivial thermodynamic limit.

At the same time, taking the partial derivative of the loss (\ref{lossLORA}) with respect to $\mA_{in}$ results in
\begin{equation}
    \label{partialAsp}
    \begin{aligned}
    \partial_{\mA_{in}} \ell & = (\Delta^\ddagger)^\mu \sum_{k=1}^{K}{\vh^\ddagger_k\,g'\left( \tilde{x}_k  \right) \frac{\beta}{\sqrt{N}} \mB_{ki} \vxi_n^\mu   }.
    \end{aligned}
\end{equation}
In matrix form, if $\mA_i$ is the $i$-th row of $\mathbf{A}$, we end up with
\begin{equation}
    \label{updateA}
    \mA_i^{\mu+1} = \mA_i^{\mu} -\beta \frac{\eta_\mA}{\sqrt{N}}  (\Delta^\ddagger)^{\mu}\left(  \sum_{k=1}^{K}{ {\vh^\ddagger_k}^{\mu}g'\left(\tilde{x}_k^{\mu}\right)\mB_{ki}^{\mu}}\right)(\vxi^{\mu})^T.
\end{equation}

\subsubsection*{Differential equations}
\label{ODEsLORA}
We are now ready to recover the various differential equations tracking the evolutions of the various order parameters. In the following, when writing $\vv$, $\vh$, $\Delta$ or $\mW_p$, we implicitly refer to the quantities associated with the second task, namely $\vv^\ddagger$, $\vh^\ddagger$,  $\Delta^\ddagger$ and $\mW_p^\ddagger$, respectively.

\paragraph{B:} We can rewrite the update rule of $\mB$ (\ref{updateD}) as
\begin{equation*}
    \frac{\mB_{ki}^{\mu+1} - \mB_{ki}^{\mu}}{1/N} = -\eta_{\mB}\,\beta\,\Delta^\mu \vh_k^\mu g'\left(\tilde{x}_k^{\mu} \right)z_i^{\mu}.
\end{equation*}
Defining $\tau:=\mu/N$ and taking the thermodynamic limit
$N,\mu\to\infty$ with $\tau$ fixed, the discrete dynamics converge
to a continuous-time evolution. The corresponding differential
equation for $\mB$ is
\begin{equation*}
    \frac{\mathrm{d}\mB_{ki}}{\mathrm{d}\tau} = -\eta_\mB\,\beta\,\vh_k \bigl\langle g'\left( \tilde{x}_k\right) \, \Delta\, z_i \bigr\rangle.
\end{equation*}
In this limit, an average over the preactivation can be explicitly taken, resulting in a deterministic time evolution. As for the generalization error, we can rewrite the update as a function of the gaussian integrals given in Appendix~\ref{app:integrals_computation}:

\begin{equation}
\label{ODEforB_integral}
\begin{split}
    \frac{\mathrm{d}\mB_{ki}}{\mathrm{d}\tau} 
    &=\eta_\mB\,\beta\,\vh_k \left[ \sum_{p=1}^{M} \vv_p \bigl\langle g'\left( \tilde{x}_k\right)\, z_i \, g\left( \nu_p \right)  \bigr\rangle - \sum_{l=1}^{K}{ \vh_l\langle g'\left( \tilde{x}_k\right)\, z_i \, g\left( \tilde{x}_l \right)  \rangle } \right] \\
    &=\eta_\mB\,\beta \,\vh_k \left[ \sum_{p=1}^{M} \vv_p I_3(K+k,2K+M+i,2K+p) \right. \\
    &\qquad \qquad \quad \left. - \sum_{l=1}^{K}{ \vh_l I_3(K+k,2K+M+i,K+l) } \right].
\end{split}
\end{equation}

\paragraph{$\vh$:} From the loss on the second task (\ref{lossLORA}), the gradient update for $\vh$ states
\begin{equation}
    \label{updateH_LORA}
    \vh_k^{\mu+1} = \vh_k^{\mu} - \frac{\eta_\vh}{N} \Delta^\mu g(\tilde{x}_k^{\mu}).
\end{equation}
Rearranging the terms of the gradient update and taking the thermodynamic limit, the differential equation for the readout weights reads
\begin{equation}
    \label{ODEforH_integral}
    \frac{\mathrm{d} \vh_k}{\mathrm{d}\tau} = \eta_\vh\left[ \sum_{p=1}^M{ \vv_p \, I_2(2K+p,K+k) } -\sum_{l=1}^K{ \vh_l\,I_2(K+l,K+k) }\right].
\end{equation}

\paragraph{\texorpdfstring{$\mPhi$}{Phi}:} Starting from the update rule of $\mA$ (\ref{updateA}) and multiplying by $(\mA_{j}^{\mu+1})^T$ on the right, we get
\begin{equation*}
\begin{aligned}
\mA_i^{\mu+1}(\mA_j^{\mu+1})^T & =  \mA_i^{\mu}(\mA_j^{\mu})^T - \eta_\mA\,\beta\,\Delta^\mu \left(\sum_{k=1}^{K}{ \vh_k^\mu g'\left(\tilde{x}_k^{\mu}\right)\mB_{ki}^{\mu}}\right)\frac{\mA_j^{\mu}\vxi^\mu}{\sqrt{N}} \\
& \quad - \eta_\mA \,\beta\,\Delta^\mu\left(\sum_{k=1}^{K}{ \vh_k^\mu g'\left(\tilde{x}_k^{\mu}\right)\mB_{kj}^{\mu}}\right)\frac{\mA_i^{\mu}\vxi^\mu}{\sqrt{N}} \\
& \quad + \eta_\mA^2\,\beta^2\,(\Delta^{\mu})^2\left(\sum_{k=1}^{K}{ \vh_k^\mu g'\left(\tilde{x}_k^{\mu}\right)\mB_{ki}^{\mu}}\right)\left(\sum_{l=1}^{K}{ \vh_l^\mu g'\left(\tilde{x}_l^{\mu}\right)\mB_{lj}^{\mu}}\right)\frac{\|\vxi^\mu\|^2}{N}.
\end{aligned}
\end{equation*}
Rewriting this quantity as a function of the order parameters and taking the thermodynamic limit results in
\begin{equation*}
    \begin{aligned}
        \frac{\mathrm{d}\mPhi_{ij}}{\mathrm{d}\tau} &= -\eta_\mA\,\beta\,\left( \sum_{k=1}^{K}{ \vh_k \mB_{ki} \bigl\langle g'\left(\tilde{x}_k\right)\,z_j\,\Delta \bigr\rangle } \right) \\
        & \quad-\eta_\mA\,\beta\,\left( \sum_{k=1}^{K}{ \vh_k \mB_{kj} \bigl\langle g'\left(\tilde{x}_k\right)\,z_i\,\Delta \bigr\rangle } \right) \\
        & \quad + \eta_\mA^2\,\beta^2\,\left( \sum_{k,l=1}^K { \vh_k \vh_l \mB_{ki} \mB_{lj} \bigl\langle g'\left(\tilde{x}_k\right)g'\left(\tilde{x}_l\right) \Delta^2 \bigr\rangle } \right).
    \end{aligned}
\end{equation*}
Finally, expanding the definition of $\Delta$ and $\Delta^2$  and writing the quantities in terms of $I_3$ and $I_4$, we obtain a closed form solution for the differential equation
\begin{equation}
    \label{ODEforPhi_integral}
    \begin{aligned}
        \frac{\mathrm{d}\mPhi_{ij}}{\mathrm{d}\tau} &= \eta_\mA\beta\left[\sum_{k=1}^K \sum_{p=1}^M {\vh_k \vv_p \mB_{ki} I_3(K+k,2K+M+j,2K+p)} \right. \\ 
        & \left.  \qquad \qquad - \sum_{k,l=1}^K{ \vh_k \vh_l \mB_{ki} I_3(K+k,2K+M+j,K+l) }\right] \\
        & \quad +\eta_\mA\beta\left[\sum_{k=1}^K \sum_{p=1}^M{\vh_k \vv_p \mB_{kj} I_3(K+k,2K+M+i,2K+p)} \right. \\ 
        & \left. \qquad \qquad \quad - \sum_{k,l=1}^K{ \vh_k \vh_l \mB_{kj} I_3(K+k,2K+M+i,K+l) }\right] \\
        & \quad + \eta_\mA^2\beta^2\left[ \sum_{k,l=1}^K \sum_{k',l'=1}^K{\vh_k \vh_l \vh_{k'} \vh_{l'} \mB_{ki} \mB_{lj} I_4(K+k,K+l,K+k',K+l')} \right. \\
        & \left. \qquad \qquad \quad -2\sum_{k,l=1}^K \sum_{k'=1}^K \sum_{p=1}^M{ \vh_k \vh_l \vh_{k'} \vv_p \mB_{ki} \mB_{lj} I_4(K+k,K+l,K+k',2K+p) } \right. \\
        & \left. \qquad \qquad \quad + \sum_{k,l=1}^K \sum_{p,q=1}^M { \vh_k \vh_l \vv_p \vv_q \mB_{ki} \mB_{lj} I_4(K+k,K+l,2K+p,2K+q) }
        \right].
    \end{aligned}
\end{equation}

    \paragraph{\texorpdfstring{$\mXi$}{Xi}:} Starting from the update rule of $\mA^T$ (\ref{updateA}),multiplying by $\mJ_{k}^{\mu+1}$ on the left and substituting the definition of $\mXi$, we get
\begin{equation*}
    \frac{\mXi_{ki}^{\mu+1}-\mXi_{ki}^{\mu}}{1/N} = - \eta_\mA \beta \Delta^\mu \left( \sum_{l=1}^K { \vh_l^\mu \,g'(\tilde{x}_l^{\mu}) \mB_{li}^{\mu}\,x_k } \right)
\end{equation*}
which can be rewritten in the thermodynamic limit as:
\begin{equation}
    \label{ODEforXi_integral}
    \begin{aligned}
    \frac{\mathrm{d}\mXi_{ki}}{\mathrm{d}\tau} &= \eta_\mA\,\beta\,\left[ \sum_{l=1}^K \sum_{p=1}^M{ \vh_l \vv_p \mB_{li} I_3(K+l,k,2K+p) } \right. \\
    & \left. \qquad \qquad - \sum_{l,k'=1}^K{ \vh_l \vh_{k'} \mB_{li} I_3(K+l,k,K+k') } \right].
    \end{aligned}
\end{equation}

\paragraph{\texorpdfstring{$\mGamma$}{Gamma}:} Starting from the update rule of $\mA^T$ (\ref{updateA}),multiplying by $\mW_{p}^{\ddagger}$ on the left and substituting the definition of $\mGamma$, we get
\begin{equation*}
    \frac{\mGamma_{pi}^{\mu+1}-\mGamma_{pi}^{\mu}}{1/N} = - \eta_\mA \beta \Delta^\mu \left( \sum_{k=1}^K { \vh_k^\mu \,g'(\tilde{x}_k^{\mu}) \mB_{ki}^{\mu} \, \nu_p } \right)
\end{equation*}
which, in the thermodynamic limit, becomes
\begin{equation}
    \label{ODEforGamma_integral}
    \begin{aligned}
    \frac{\mathrm{d}\mGamma_{pi}}{\mathrm{d}\tau}  &= \eta_\mA\,\beta\,\left[ \sum_{k=1}^K \sum_{q=1}^M{ \vh_k \vv_q \mB_{ki} I_3(K+k,2K+p,2K+q) } \right. \\
    & \left. \qquad \qquad- \sum_{k,l=1}^K { \vh_k \vh_l \mB_{ki} I_3(K+k,2K+p,K+l) } \right].
    \end{aligned}
\end{equation}

\paragraph{\texorpdfstring{$\mLambda$}{Lambda}:} Starting from the update rule of $\mA^T$ (\ref{updateA}), multiplying by $\mW_{p}^{\dagger}$ on the left and substituting the definition of $\mLambda$, we get
\begin{equation*}
    \frac{\mLambda_{mi}^{\mu+1}-\mLambda_{mi}^{\mu}}{1/N} = - \eta_\mA \beta \Delta^\mu \left( \sum_{k=1}^K { \vh_k^\mu \,g'(\tilde{x}_k^{\mu}) \mB_{ki}^{\mu} \, \rho_m } \right)
\end{equation*}
which becomes in the thermodynamic limit:
\begin{equation}
    \label{ODEforLambda_integral}
    \begin{aligned}
    \frac{\mathrm{d}\mLambda_{mi}}{\mathrm{d}\tau} & = \eta_\mA\,\beta\,\left[ \sum_{k=1}^K \sum_{p=1}^M{ \vh_k \vv_p \mB_{ki} I_3(K+k,2K+M+L+m,2K+p) } \right. \\
    & \left. \qquad \qquad - \sum_{k,l=1}^K{ \vh_k \vh_l \mB_{ki} I_3(K+k,2K+M+L+m,K+l) } \right].
    \end{aligned}
\end{equation}

\section{Additional results concerning the SDGM}
\label{sec:appendix_SDGM}

\subsection{SDGM Implementation Details}\label{sec:SDGM_implementation}

In the teacher-student model, the multi-head architecture provides a particularly simple measure of feature importance. After Task 1 training, the magnitude of the readout coefficient $|\vh_i^\dagger|$ quantifies the contribution of hidden unit $i$ to the
Task 1 prediction. Since the $i$-th readout coefficient multiplies the feature generated by the $i$-th row of the first-layer matrix, hidden units with large $|\vh_i^\dagger|$ identify feature directions that are most strongly used by Task 1. We therefore rank the hidden units according to their Task 1 readout magnitudes and protect the most
important ones during Task 2 adaptation.

Formally, for a given \(\kappa\leq K\), let \(\mathcal{S}_{\mathrm{frozen}}\subset\{1,\ldots,K\}\) denote the indices corresponding to the \(\kappa\) largest values of \(|\vh^\dagger|\), so that \(|\mathcal{S}_{\mathrm{frozen}}|=\kappa\).
 The complementary set, $\mathcal{S}_{\mathrm{plastic}}
    =
    \{1,\ldots,K\}\setminus\mathcal{S}_{\mathrm{frozen}}
$, contains the hidden units available for Task 2 adaptation. Importantly, the partition is determined only after Task 1 has been learned and therefore depends on the state reached by the network at the task switch, rather than on a fixed architectural partition specified before training.

We implement SDGM directly within the LoRA parameterization. Rather than optimizing both LoRA factors, we fix the up-projection matrix $\mB$ to a sparse matrix $\bm\Omega\in\{0,1\}^{K\times L}$ whose non-zero rows are restricted to $\mathcal{S}_{\mathrm{plastic}}$, and optimize only $\mA$ and $\vh^\ddagger$. Let $u_1<\cdots<u_{K-\kappa}$ denote the elements of $\mathcal{S}_{\mathrm{plastic}}$. We assign each plastic unit to one adapter direction, cycling through the $L$ directions, that is
\begin{equation}
    \bm\Omega_{ij} =
    \begin{cases}
        1, & \text{if } i=u_k \text{ and } j=1+((k-1)\bmod L) \text{ for some } k\in\{1,\ldots,K-\kappa\},\\
        0, & \text{otherwise}.
    \end{cases}
    \label{eq:sdgm_mask}
\end{equation}
Since $\bm\Omega$ has nonzero rows only on the $K-\kappa$ plastic units, the update $\mB\mA$ has rank at most $\min(K-\kappa, L)$. For $L < K-\kappa$, several plastic units share the same adapter direction while for $L > K-\kappa$, the last $L-(K-\kappa)$ columns of $\bm\Omega$ vanish and the corresponding rows of $\mA$ are never updated. Throughout the main text we set $\kappa = K - L$, so that the update has rank exactly $L$ and LoRA and LoRA + SDGM are compared at equal rank. The rank sweep at fixed $\kappa$ is reported in Appendix.~\ref{sec:fixed_kappa}.

Because $\bm\Omega$ is built from the Task 1 state and then held fixed, the same macroscopic theory built for LoRA applies by setting $\mB= \bm\Omega$ and $\mathrm{d}\mB / \mathrm{d} \tau = 0$ in Eq.~\ref{ODEforB_integral}, while integrating Eqs.~\ref{ODEforPhi_integral}--\ref{ODEforLambda_integral}. The rows of the update associated with frozen units therefore remain zero during Task 2. Since $\vh^\dagger$ is a finite-dimensional parameter tracked explicitly by the theory, the partition $\mathcal{S}_{\mathrm{frozen}}$ is itself predicted by the Task 1 ODEs.
\paragraph{Comparison with standard + SDGM.}
For completeness, we also apply the same state-dependent partition to standard fine-tuning. This comparison can be represented within the same parameterization. By taking $L=K$, setting $\gamma=\sqrt{K}$ and $\mB=\mI_K$, we get $\mJ=\mJ_{s}+\mA.$ With $\mA$ initialized at zero, optimizing $\mA$ is equivalent to updating $\mJ$ directly from $\mJ_s$: standard sequential training is thus totally contained as a sub-case of LoRA fine-tuning. Moreover, replacing $\mI_K$ by the corresponding diagonal SDGM mask therefore yields masked full fine-tuning as a special case of the same framework.

\subsection{Applying the inverse and random SDGM}
\label{app:inv_SDGM}
\begin{figure}[h!]
    \centering
    \includegraphics[width=0.8\linewidth]{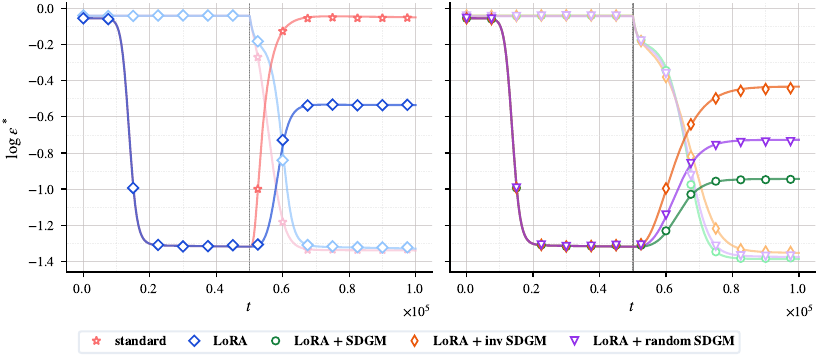}
    \caption{\textbf{Ablation of the selection protocol via Inverse and Random SDGM.} Generalization error trajectories on Task 1 and Task 2 under various selection protocols. Left: standard fine-tuning and vanilla LoRA, where $\mB$ is trained. Right: LoRA with $\mB$ fixed to a mask that freezes the $\kappa$ units with the largest (SDGM), smallest (inverse SDGM) or randomly chosen (random SDGM) Task 1 readouts. Comparing the methods disentangles the effect of fixing $\mB$ from that of targeted feature protection. Parameters: $N =10^3$, $K = 10$, $M = 5$, $L = 5$, $c = 0.5$, $\alpha = 50$, $\kappa = 5$.}
    \label{fig:inverse_scoring}
\end{figure}

SDGM differs from vanilla LoRA in two ways: the up-projection $\boldsymbol{B}$ is fixed to a mask $\boldsymbol{\Omega}$ instead of being trained, and the mask is built from the Task 1 state. To verify that the second ingredient, feature selection, drives SDGM performance rather than the restriction of the trainable subspace alone, we perform two ablation experiments. In the \textit{Inverse SDGM} protocol, $\mathcal{S}_{\mathrm{frozen}}$ isolates the $\kappa$ smallest magnitude entries of $\vh^\dagger$, freezing the least informative directions relative to Task 1. In the \textit{Random SDGM} protocol, $\mathcal{S}_{\mathrm{frozen}}$ is a set of $\kappa$ hidden units drawn uniformly at random, independently of $\vh^\dagger$. 
SDGM, Inverse SDGM and Random SDGM all fix $\boldsymbol{B}$ to a mask with the same number of frozen units at the same adapter rank. They therefore impose identical capacity constraints, and any difference between them comes from which units are protected. Comparing them with vanilla LoRA, where $\boldsymbol{B}$ is trained, then separates the effect of fixing $\boldsymbol{B}$ from that of selecting the protected units.

Fig.~\ref{fig:inverse_scoring} shows that all protocols reach comparable Task 2 performance, while their Task 1 retention is clearly ordered: standard fine-tuning forgets the most, followed by Inverse SDGM, vanilla LoRA, Random SDGM and SDGM. Fixing $\boldsymbol{B}$ alone therefore does not explain the gain. Inverse SDGM also fixes $\boldsymbol{B}$, yet it forgets more than vanilla LoRA, because freezing the units least important for Task 1 concentrates the whole update on the units that carry it. What matters is the choice of the protected units: Random SDGM already forgets less than vanilla LoRA, and SDGM, which protects the units with the largest Task 1 readouts, reduces forgetting further.

Random selection thus serves as the reference at equal capacity: protecting the units with the largest Task 1 readouts improves on it, while protecting the least informative ones makes forgetting worse than not masking at all. This shows that effective feature protection requires explicitly identifying and freezing key task-relevant representations, rather than merely fixing the up-projection or reducing the dimensionality of the trainable subspace.

\subsection{Validation under an unbounded activation function}
    \label{app:ReLU_trial}

\begin{figure}[h!]
    \centering
    \includegraphics[width=0.8\linewidth]{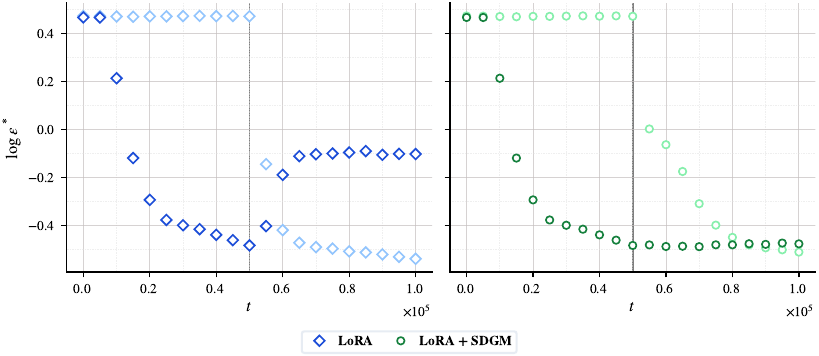}
    \caption{\textbf{Validation of SDGM under unbounded ReLU activations.} Generalization error dynamics on Task 1 and Task 2 when both teacher and student networks employ ReLU as the activation function. The lower error on Task 1 confirms that the proposed selection rule mitigates catastrophic forgetting independently of activation saturation. Parameters: $N=10^3$, $K=10$, $M=5$, $L=5$, $c=0.5$, $\alpha = 50$. Results shown here are experiment-only.}
    \label{fig:unbounded_act_func}
\end{figure}

To verify that the efficacy of SDGM comes from structural information routing rather than artifacts of activation saturation, we evaluate the protocol under an unbounded activation function. Smooth, bounded activations such as $\mathrm{erf}(z)$ naturally constrain preactivation magnitudes. In contrast, the Rectified Linear Unit (ReLU), defined as $g(z) = \max(0, z)$, exhibits unbounded values after the first layer. As illustrated in Fig.~\ref{fig:unbounded_act_func}, applying the proposed selection protocol under ReLU dynamics successfully preserves Task 1 performance throughout Task 2 adaptation, with similar transfer on Task 2. This suggests that the protocol does not merely exploit head specialization but actively isolates and protects the sub-network carrying critical task representations.

\subsection{Rank sweep at a fixed number of frozen units}\label{sec:fixed_kappa}

In the main text, we set $\kappa = K - L$, so that varying $L$ in Fig.~\ref{fig:forgetting+transfer}c changes both the adapter rank and the number of frozen units. To separate the two effects, we repeat the rank sweep of LoRA + SDGM at fixed $\kappa \in \{2, 5, 8\}$, for $L = 1, \dots, 10$, both in the unspecialized regime of the main text and in the specialized regime of Appendix~\ref{app:specialization}. Each curve shows forgetting against transfer during Task 2 training. Since time does not appear explicitly in this representation, runs that differ only in their speed can be compared directly.

\begin{figure}[t!]
    \centering
    \begin{minipage}[t]{0.48\linewidth}
        {\raggedright \textbf{a)}\par}
        \vspace{2pt}
        \centering
        \includegraphics[width=\linewidth]{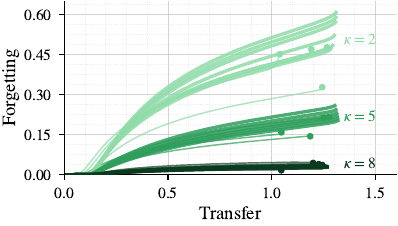}
    \end{minipage}%
    \hfill
    \begin{minipage}[t]{0.48\linewidth}
        {\raggedright \textbf{b)}\par}
        \vspace{2pt}
        \centering
        \includegraphics[width=\linewidth]{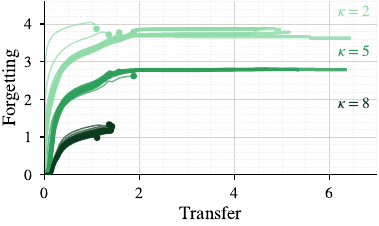}
    \end{minipage}
    \caption{\textbf{Rank sweep of LoRA + SDGM at a fixed number of frozen units.} Forgetting versus transfer during Task 2 training for $\kappa = 2, 5, 8$ and $L = 1, \dots, 10$, in \textit{a)} the unspecialized regime of the main text and \textit{b)} the specialized regime of Appendix~\ref{app:specialization}. Thick lines correspond to $L \geq M$, thin lines ending with a dot to $L < M$ (the dot marks the end of training in \textit{a)} and the Task 2 plateau in \textit{b)}). We average over 10 different seeds, at equal time in \textit{a)} and at equal transfer in \textit{b)}, where runs with different $L$ progress at different speeds due to the LoRA prefactor $\gamma / \sqrt{L}$. Parameters: $N = 10^3$, $K = 10$, $M = 5$, $c = 0.5$; \textit{a)} $\alpha = 50$; \textit{b)} $\alpha = 2000$ for Task 1 and $\alpha = 8000$ for Task 2.}
    \label{fig:fixed_kappa}
\end{figure}

Fig.~\ref{fig:fixed_kappa}a shows the unspecialized regime. Increasing the number of protected units reduces forgetting, at little cost in transfer, for every rank $L$. A large $\kappa$ retains most of the transfer because the student has not yet specialized to the individual teacher features at the training horizon considered, and the Task 2 error it reaches can be attained with few plastic units.

The specialized regime exposed in Fig.~\ref{fig:fixed_kappa}b reveals qualitative differences with respect to the unspecialized one. Representing Task 2 now requires on the order of $M$ specialized units, and the effective rank $\min(L, K-\kappa)$ of the update becomes the limiting factor. LoRA + SDGM escapes the Task 2 plateau only if $\min(L, K-\kappa) \geq M$. For $L < M$, or for $\kappa = 8$ at any rank, Task 2 stays on the plateau: in this regime, too large a $\kappa$ does limit transfer. Once the plateau is escaped, increasing $L$ beyond $K-\kappa$ only adds vanishing columns to $\bm\Omega$ (Eq.~\ref{eq:sdgm_mask}), and its sole effect is to slow down the dynamics through the LoRA prefactor $\gamma/\sqrt{L}$, so that the curves for different $L$ follow nearly the same path in the forgetting--transfer plane.

In summary, $\kappa$ sets the forgetting at a given transfer in both regimes, while in the specialized regime the effective rank $\min(L, K-\kappa)$ also determines whether Task 2 can be learned at all. Fixing $\kappa$ independently of $L$, however, requires an assumption on how many units carry the Task 1 features, which the choice $\kappa = K - L$ of the main text avoids.

\section{Results in the specialized regime}
\label{app:specialization}
All the results presented in the main text are shown in the so-called \textit{symmetric regime}, where the student has not yet been able to specialize towards the specific directions of the teacher. The motivation for this choice of regime is multiple. First, it is more difficult to align with the first task in the \textit{overrealizable regime}, that is when $K>M$. Second, the time constant associated with symmetric subspace escape increases linearly with $K$. A full discussion on this problem can be found in \citet{PhysRevE.52.4225}. Finally, with our choice of readout initialization, the specialization is even more difficult. Multiple works have been done to understand the impact of initialization on forgetting in an equivalent setting, as well as proposing good habits for the initialization scheme \citep{lee2022,jarvis2025}. However, these good habits can be applied with \textit{a priori} knowledge on the tasks that must be fitted, a setting very different from practitioners' experience.\\
We present here additional results in the \textit{specialized regime}.
Following the insights on initialization from \citet{jarvis2025}, we set 
\begin{align*}
    \vh^\dagger_i = 
    \begin{cases}
        10^{-2} & \text{if } 1 \le i \le \lfloor K/2 \rfloor, \\
        0 & \text{otherwise},
    \end{cases}
    \qquad \text{and} \qquad
    \vh^\ddagger_i = 
    \begin{cases}
        -1 & \text{if } 1 \le i \le \lfloor K/2 \rfloor, \\
        0 & \text{otherwise}.
    \end{cases}
\end{align*}

\begin{figure}[t!]
    \centering
    \begin{minipage}[t]{0.73\linewidth}
        {\raggedright \textbf{a)}\par}
        \vspace{2pt}
        \centering
        \includegraphics[width=\linewidth]{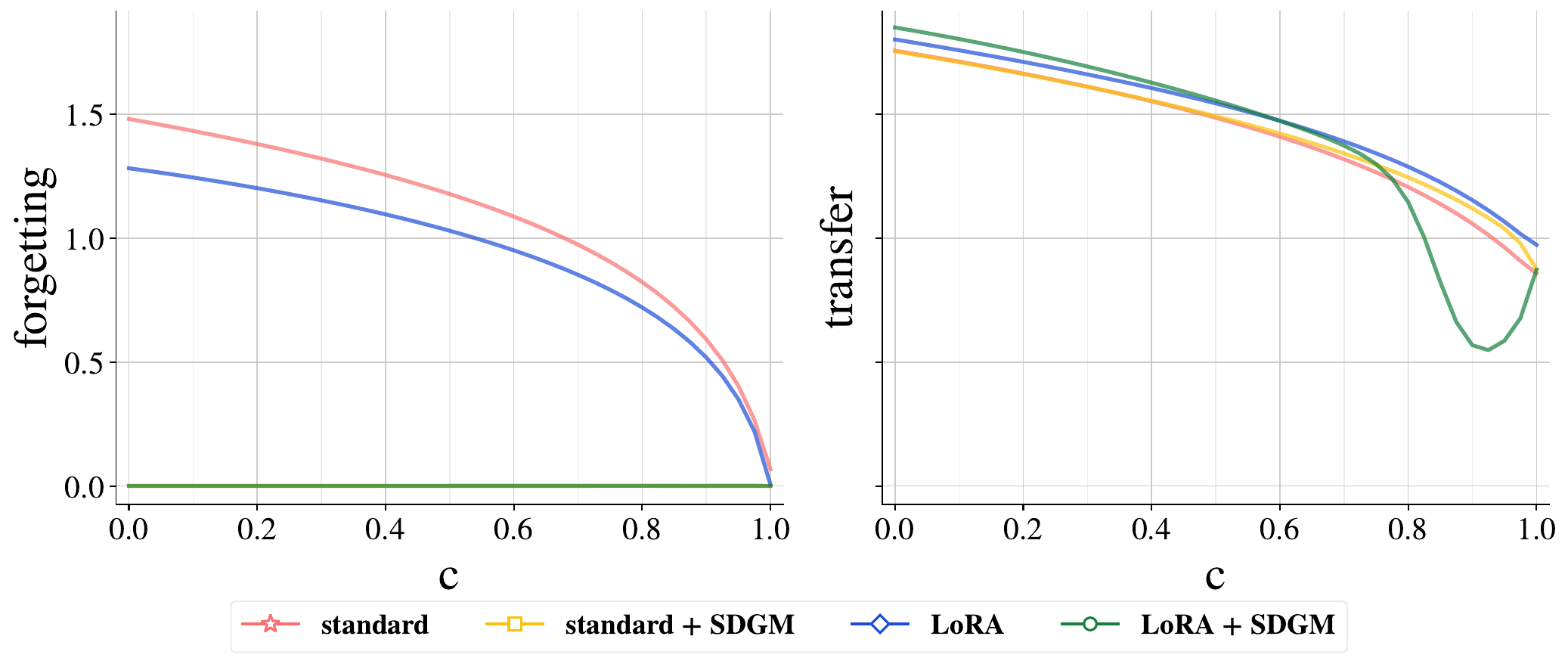}
        \label{fig:forgetting_unspecialized}
    \end{minipage}%
    \hfill
    \begin{minipage}[t]{0.25\linewidth}
        {\raggedright \textbf{b)}\par}
        \vspace{2pt}
        \centering
        \includegraphics[width=\linewidth]{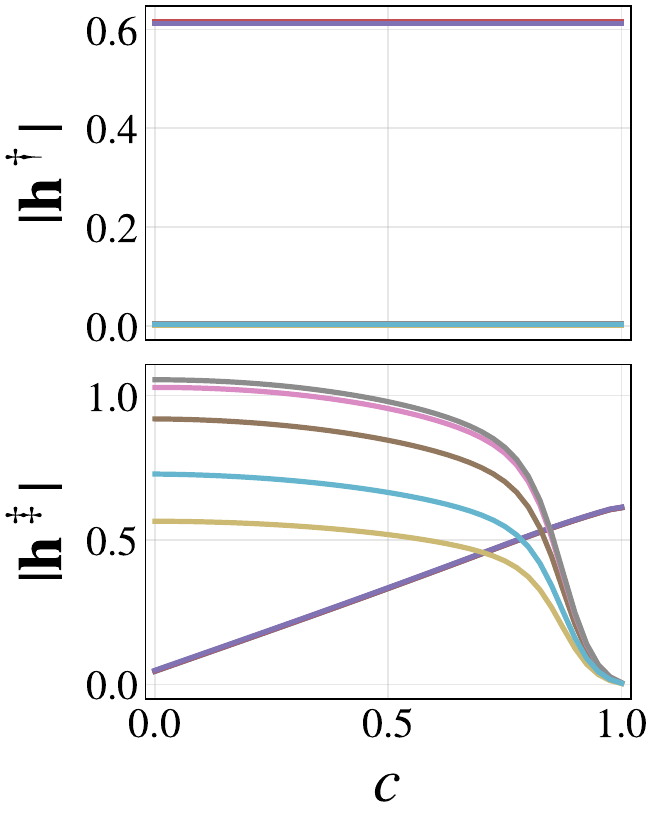}
    \end{minipage}
    \hfill
    \caption{\textbf{ Typical forgetting on the first task and transfer on the second task with the new initialization}. In this setting, the SDGM procedure allows for no forgetting on the full range of task similarity, while allowing for the same transfer. The smaller transfer at big task similarity for LoRA + SDGM is due to long symmetric plateau, whose size increases non-monotonically with $c$. Parameters : $N=10^3$, $K=10$, $M=5$, $L=5$, $\alpha=50$.}
    \label{fig:heads_unspecialized_regime}
\end{figure}

We first check the impact of this new initialization on the readout weights in the unspecialized case in Fig.~\ref{fig:heads_unspecialized_regime}. By artificially forcing the network to only use a fraction of its directions to learn Task 1, SDGM allows for no forgetting on the full range of task similarity. This can be understood by checking the Task 1 readout weights and uncovering that only a fraction of their value is non-0: the initialization biases the network dynamics towards self-pruning, letting free directions for Task 2.  At the same time, when training on Task 2, the magnitude of the readout weights associated to new direction decreases monotonically with task similarity. This effect is a consequence of \textit{node re-use} \citep{lee2022}, where the student is able to recycle directions learned on Task 1, already partially aligned with Task 2. Eventually, when $c=1$, the student is not learning any new directions, even after training.\\

\begin{figure*}[h!]
    \includegraphics[width=0.95\linewidth]{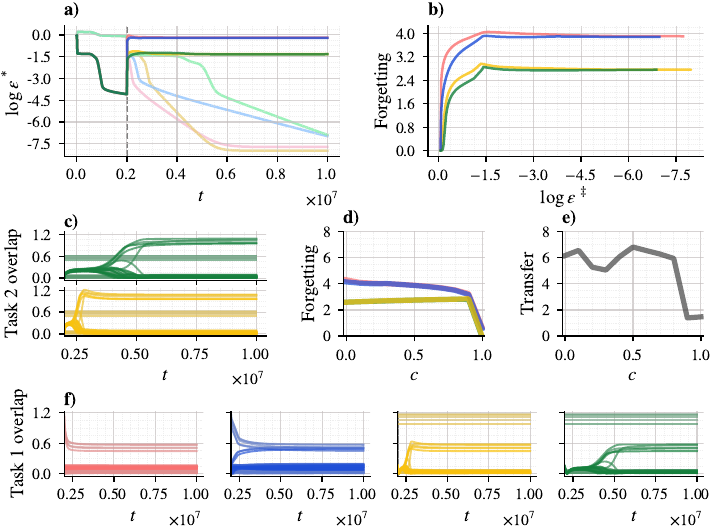}
    \caption{\textbf{Additional results in the specialized regime.} \textit{a)} Typical generalization error for full fine-tuning (red), LoRA (blue), and their SDGM-constrained variants (yellow and green). During Task 1 training, the symmetric plateau, corresponding to the absence of specialization, is located at $\log \epsilon^* \approx -1.5$. \textit{b)} Typical forgetting behavior as a function of the Task 2 generalization error. SDGM decreases the forgetting by protecting important features learned on Task 1, even in the specialized regime. 
    \textit{c)} Student overlaps with the Task 2 teacher, as defined in Eq.~\ref{eq:tilde_x_nu} after the task switch for LoRA + SDGM and standard + SDGM. The overlaps corresponding to frozen student directions remain constant throughout Task 2 training. \textit{d)} Task 1 forgetting and \textit{e)} Task 2 transfer as a function of teacher similarity $c$. For a fair comparison, the forgetting presented is shown at equal transfer between all methods for all values of $c$. 
    \textit{f)} Student overlaps with the Task 1 teacher, as defined in Eq.~\ref{eq:tilde_x_rho} after the task switch for the various procedures. In this regime, standard fine-tuning and LoRA overwrite the useful directions for Task 1, which are preserved by the SDGM procedure. Results shown here are theory-only.}

    \label{fig:app_specialization_full_scheme}
\end{figure*}

We now turn to the specialized regime, presented in Fig.~\ref{fig:app_specialization_full_scheme}. With this initialization, Task 1 reaches the specialized phase within $\alpha = 2000$, but Task 2 requires a much longer horizon due to LoRA with SDGM being stuck on a long symmetric plateau. We therefore extend Task 2 training to $\alpha = 8000$, so that all methods reach the exponential decrease in generalization error characteristic of specialization, as observed in Fig.~\ref{fig:app_specialization_full_scheme}a for $c = 0.5$. As in the unspecialized regime, LoRA and its SDGM-constrained variant adapt more slowly to Task 2. The plateau is longest for LoRA with SDGM, since its plastic hidden units must be recruited from scratch while the units carrying Task 1 are frozen. Fig.~\ref{fig:app_specialization_full_scheme}c shows this mechanism: the frozen units keep their overlap with the Task 2 teacher unchanged, while the plastic units progressively align with it, much faster for standard fine-tuning with SDGM than for its LoRA counterpart due to the effective learning rate being bigger.

Because the methods learn Task 2 at different speeds, comparing forgetting at a fixed training time would conflate stability with learning speed. We therefore compare the methods at equal Task 2 performance. Fig.~\ref{fig:app_specialization_full_scheme}b reports Task 1 forgetting as a function of the Task 2 error reached along a typical training. At every level of Task 2 error, both SDGM variants forget less than standard fine-tuning and LoRA. In this regime, LoRA forgets as much as standard fine-tuning: with the node re-use initialization, both methods rewrite the hidden units specialized on Task 1 (Fig.~\ref{fig:app_specialization_full_scheme}f).

We apply the same principle across task similarities. For each $c$, we fix the transfer to the largest value reached by all four methods within the training window and measure the forgetting of each method at the first time it reaches this transfer, as shown in Figs.~\ref{fig:app_specialization_full_scheme}d--e. At matched transfer, SDGM forgets less than standard fine-tuning and LoRA over the whole range of task similarity, and both SDGM variants show the same forgetting. Unlike in the unspecialized regime, the forgetting of standard fine-tuning and LoRA decreases with $c$: the re-used units end up aligned with the Task 2 teacher, whose overlap with the Task 1 teacher is $c$ \citep{lee2022}.

\section{Additional details on LoRA's initialization}
\label{app:LoRA}
In this controlled continual learning setting, the initialization of the LoRA matrices demands careful consideration. The foundational principle of LoRA is to ensure that the weight perturbation is equal to 0 at initialization, that is we force 
$\Delta \mJ = 0$ when adding the LoRA adapter in order to prevent an immediate disruption of the parameter  configuration at the task switch. While the initialization scheme proposed in the  seminal LoRA framework \citep{hu2022lora} is tailored to maximize downstream task performance and training stability by letting $\mA$ start from a Kaiming initialization and setting $\mB=0$, our objective introduces a distinct trade-off: we want to achieve high plasticity on Task 2 while keeping stability on Task 1.

Thus, another possible initialization scheme, recently proposed in \citet{2604.01694} is to inverse this choice and to let $\mA = 0$, $\mB = \mathcal{O}(1)$. This option demonstrated comparable or superior performance across a variety of downstream tasks while mitigating forgetting on Task 1. This mitigation can be understood by first looking at the classical initialization mechanism: initial gradient with respect to $\mA$ vanishes, leaving the early updates to be driven entirely by the evolution of $\mB$. In this regime, the random weights of $\mA$ act as a static random feature projector. This random projection disrupts the alignment between $\mJ_{s}$ and Task 1. Consequently, the optimization trajectory on Task 2 drives the system into a regime of catastrophic forgetting. 

Conversely, the new initialization prevents this destructive mechanism.
In this case, the gradient updates of $\mB$ vanish, forcing $\mA$ to absorb the initial learning dynamics. The adaptation thus propagates through the low-rank bottleneck $L$ in a more constrained manner, allowing the network to selectively acquire features relevant to Task 2 while maintaining minimal structural overlap with the representation learned for Task 1.

At the same time, randomly selecting the rows of $\mB$ makes the optimization landscape highly sensitive to initialization, resulting in substantial variability in the trajectory and final configuration of $\mA$ across random seeds. Within the proposed theoretical framework, this sensitivity is directly visible in the overlaps we recover: since $\mB$ acts as an order parameter of the system, its initial configuration has a strong impact on the overall training dynamics. To reduce this run-to-run variability, we \textit{initialize} $\mB$ deterministically as

$$
\mB_{k\ell}
=
\mathbf{1}\!\left\{\ell = 1 + ((k-1)\bmod L)\right\}.
$$

This does not alter LoRA's parameterization, for both $\mB$ and $\mA$ are trainable low-rank adapters and yields reproducible initial conditions for the corresponding ODE dynamics.

\section{Deep multi-task benchmark}
\label{app:deep_benchmark}

The theory of the main text is derived for a two-layer network trained online on two tasks. To test whether its mechanisms survive outside the solvable model, we apply the same procedures to a deep convolutional network trained on a sequence of six tasks of natural images. More precisely, we want to check the ordering of the methods at matched transfer, and the quantities through which the theory explains forgetting.

We use Split CIFAR-100 \citep{Krizhevsky09} with a fixed class order. Task 1 contains 50 classes, and each of the five following tasks contains 10 classes. As in the teacher-student model, the setting is task-incremental, each task has its own linear readout head, and the heads of previous tasks are frozen once their task is learned. Forgetting is therefore caused by changes in the shared layers alone.

To learn this sequence of tasks, we use a ResNet-18 \citep{he2015} with the usual CIFAR stem, i.e.\ a $3\times3$ first convolution and no max-pooling. Task 1 is learned from scratch with standard SGD for 60 epochs, with a one-cycle learning-rate schedule of maximal value 0.1, weight decay $5\times10^{-4}$, random crops and horizontal flips. 
The resulting network plays the role of the Task 1 representation $\mJ_s$ of the main text, and one such checkpoint per seed is shared by all methods. Tasks 2 to 6 are each learned for 20 epochs with SGD (momentum 0.9, no weight decay, batch size 128) and a cosine learning-rate schedule within each task. These settings are identical for all methods, only the method and its learning rate change. 

We compare the four procedures of the main text. In standard fine-tuning, all $3\times3$ convolutions are trained. In LoRA, each $3\times3$ convolution with weights viewed as a matrix $\mW_s$  is adapted as $\mW = \mW_s + \frac{\gamma}{\sqrt{L}} \mB \mA$, with rank $L = 8$ and $\gamma = 1$, as in the main text. 
The factors are initialized as in Appendix~\ref{app:LoRA}, with $\mA = 0$ and $\mB$ deterministic and merged into the weights at the end of each task. They are replaced by a fresh adapter for the next one. In standard + SDGM, the protected output channels of each convolution receive no update. In LoRA + SDGM, $\mB$ is fixed to the mask $\bm\Omega$ of Eq.~(\ref{eq:sdgm_mask}), restricted to the plastic channels, and only $\mA$ and the readout head are trained.

The SDGM strategy must be adapted to a deep network. The hidden units of the two-layer model become the output channels of each convolution: as in the main text, row $i$ of $\mB$ controls output channel $i$. The importance of a channel, the deep analogue of the readout magnitude $|\vh^\dagger_i|$, is measured by a first-order Taylor score
\begin{equation}
    s_i = \mathbb{E}\left| \sum_{h,w} a_{i,hw} \, \frac{\partial \epsilon}{\partial a_{i,hw}} \right|,
\end{equation}
where $a_{i,hw}$ is the output of channel $i$ at spatial position $(h, w)$ and $\epsilon$ is the loss of the task just learned, evaluated with its own head \citep{molchanov2017}. After Task 1, the half of the channels of each layer with the largest score are protected, which corresponds to $\kappa = K/2$ as in the main text. After each later task, an additional 5\% of the channels of each layer are protected, chosen among those still plastic. Standard + SDGM and LoRA + SDGM use the same protected sets.

The transfer and forgetting metrics must be slightly revised for sequential tasks. Let $\epsilon_j(t)$ be the test loss of task $j$ at training step $t$, and $t_i$ the step at which training on task $i$ starts. While task $i \geq 2$ is learned, we define
\begin{equation}
    \mathrm{Transfer}_i(t) = \log \epsilon_i(t_i) - \log \epsilon_i(t), \qquad
    \mathrm{Forgetting}_i(t) = \frac{1}{i-1} \sum_{j < i} \big[ \log \epsilon_j(t) - \log \epsilon_j(t_i) \big].
\end{equation}
$\mathrm{Forgetting}_i(t)$ is thus the average forgetting of all tasks learned before task $i$, measured from the start of training on task $i$. The two-task protocol of the theory then corresponds to $i = 2$. For the whole sequence, the curves of Tasks 2 to 6 are averaged on a common transfer axis. Since the methods learn at different speeds and with different learning rates, we compare them at matched transfer: forgetting is plotted against transfer along training, and read at the largest transfer reached by all methods such that a method cannot appear more stable simply because it learns less. Each method is run over a grid of learning rates 
with three seeds each. 
We also report the final average accuracy over the six tasks, the plasticity, i.e.\ the accuracy on each of Tasks 2 to 6 right after learning it, averaged over tasks, and the backward transfer (BWT), i.e.\ the mean change in accuracy of each previous task between the end of its training and the end of the sequence.

To connect the deep results to the theory, we measure two deep analogues of its order parameters. The first is the share of the update that falls on the channels most important for Task 1,
\begin{equation}
    \rho = \frac{\sum_{\ell} \sum_{i \in \mathcal{S}^{(\ell)}_\mathrm{imp}} \| \Delta \mW^{(\ell)}_i \|^2}{\sum_{\ell} \| \Delta \mW^{(\ell)} \|^2},
\end{equation}
where $\Delta \mW^{(\ell)}$ is the change of the weights of convolution $\ell$ after learning Task 2, $\Delta \mW^{(\ell)}_i$ its row for output channel $i$, and $\mathcal{S}^{(\ell)}_\mathrm{imp}$ the half of the channels with the largest Task 1 score, i.e.\ the set protected by SDGM. 
Since half of the channels are labeled important, an update that ignores importance gives $\rho = 1/2$. This quantity plays the role of the rows of $\mB$ in the change of the Task 1 alignment $\beta \mB \bm\Lambda^\top$ of Eq.~\ref{eq:tilde_x_rho}. The second is the similarity of the Task 1 representation to its state at the first task switch, measured with linear centered kernel alignment (CKA; \citet{kornblith2019}),
\begin{equation}
    \mathrm{CKA}(\mX, \mY) = \frac{\| \mY^\top \mX \|_F^2}{\| \mX^\top \mX \|_F \, \| \mY^\top \mY \|_F},
\end{equation}
where the rows of the centered matrices $\mX$ and $\mY$ are the features of the Task 1 test images at the first switch and at the current step. 
CKA is invariant to orthogonal transformations and isotropic scaling of the features, and equals 1 when the two representations coincide up to such transformations. It is the deep analogue of the overlaps of the student with the Task 1 teacher presented in Fig.~\ref{fig1:StandardVSLoRA}c.

\begin{figure}[t!]
    \centering
    \includegraphics[width=0.8\linewidth]{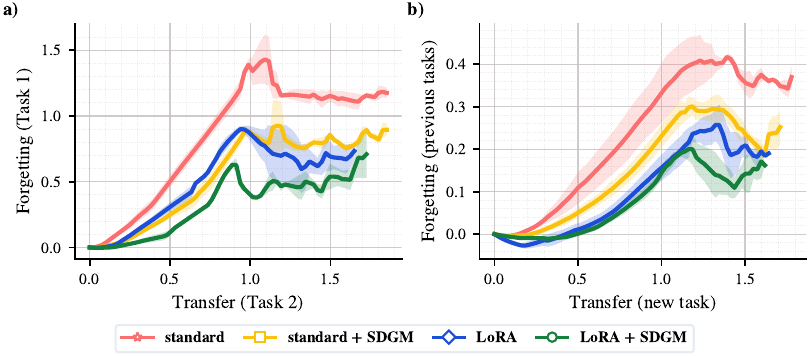}\caption{\textbf{Forgetting at matched transfer in a deep multi-task setting.} ResNet-18 on Split CIFAR-100 (six tasks). \textit{a)} Task 1 forgetting against transfer on Task 2. \textit{b)} Forgetting of all previous tasks against transfer on the new task, averaged over Tasks 2 to 6. Shaded regions show the standard deviation over three seeds. For LoRA, $L=8$.}
    \label{fig:deep_frontier}
\end{figure}

\begin{table}[t!]
    \centering
    \begin{tabular}{lccc}
        \toprule
        Method & Final avg.\ accuracy (\%) & Plasticity (\%) & BWT (\%) \\
        \midrule
        Standard & $74.5 \pm 1.4$ & $89.1 \pm 0.1$ & $-15.9 \pm 1.6$ \\
        Standard + SDGM & $78.3 \pm 0.3$ & $84.9 \pm 0.2$ & $-7.2 \pm 0.5$ \\
        LoRA ($L=8$) & $75.2 \pm 0.9$ & $85.9 \pm 0.1$ & $-11.9 \pm 1.2$ \\
        LoRA + SDGM ($L=8$) & $78.7 \pm 0.5$ & $84.3 \pm 0.4$ & $-6.1 \pm 0.4$ \\
        \bottomrule
    \end{tabular}
    \caption{\textbf{Accuracies on Split CIFAR-100 after six tasks}, at the best learning rate of each method (mean $\pm$ std over three seeds).}
    \label{tab:deep_acc}
\end{table}

\begin{figure}[t!]
    \centering
    \begin{minipage}[t]{0.48\linewidth}
        {\raggedright \textbf{a)}\par}
        \vspace{2pt}
        \centering
        \includegraphics[width=\linewidth]{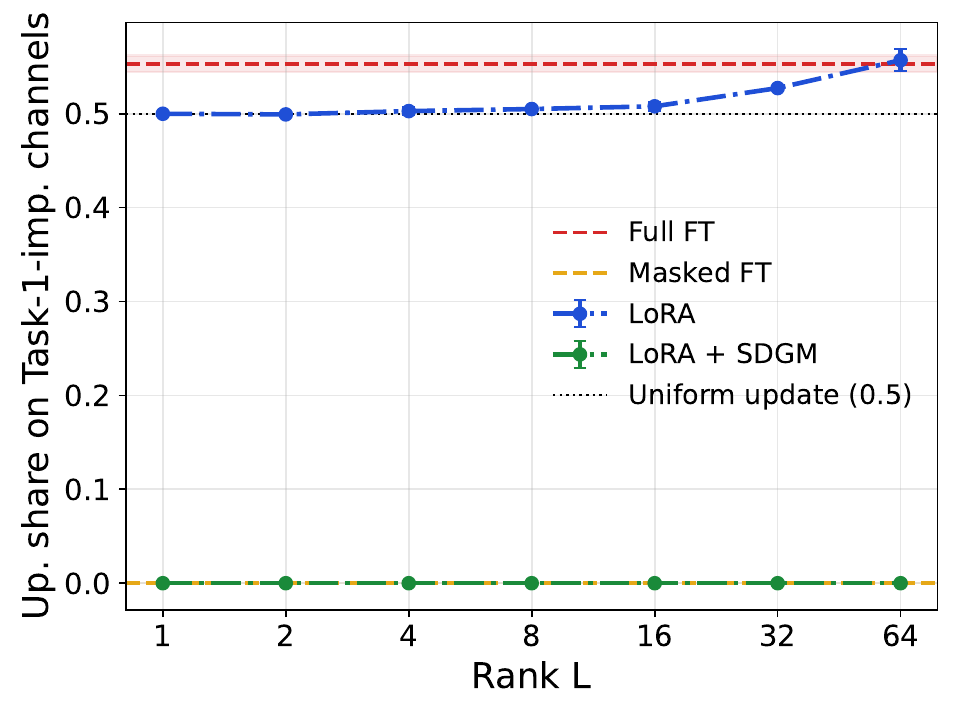}
    \end{minipage}%
    \hfill
    \begin{minipage}[t]{0.48\linewidth}
        {\raggedright \textbf{b)}\par}
        \vspace{2pt}
        \centering
        \includegraphics[width=\linewidth]{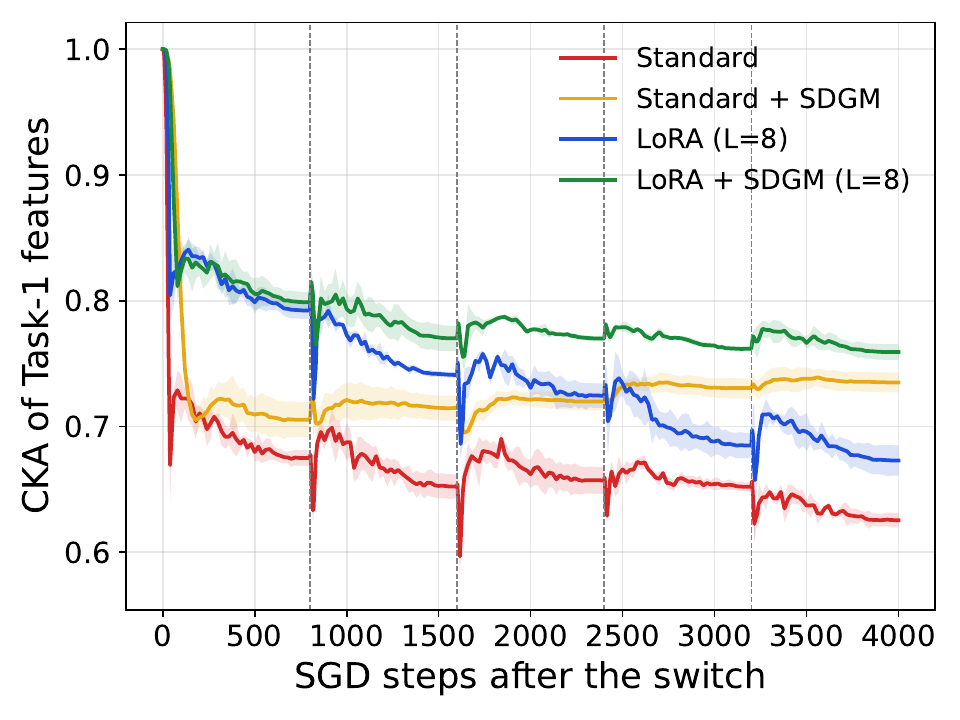}
    \end{minipage}
    \caption{\textbf{Mechanism measurements in the deep network.} \textit{a)} Share $\rho$ of the update that falls on the output channels most important for Task 1, after learning Task 2, as a function of the rank $L$. Dashed lines show standard and standard + SDGM; standard + SDGM coincides with LoRA + SDGM at zero. The dotted line marks the chance level $\rho = 1/2$. \textit{b)} Linear CKA between the Task 1 features during training and at the first task switch, over Tasks 2 to 6 ($L = 8$). Dotted vertical lines mark the task switches; each method is shown at its best learning rate. Shaded regions and error bars show the standard deviation over three seeds.}
    \label{fig:deep_share}
\end{figure}

Fig.~\ref{fig:deep_frontier} shows forgetting against transfer for the four methods. The ordering predicted by the theory carries over: at matched transfer, standard fine-tuning forgets the most, standard + SDGM and LoRA forget less, and LoRA + SDGM forgets the least, both in the two-task protocol and averaged over the six-task sequence. The relative position of LoRA and standard + SDGM differs from the unspecialized regime of Fig.~\ref{fig:protocol}a and depends on the setting; the position of LoRA + SDGM does not. For direct comparison, Fig.~\ref{fig:deep_frontier}a is the same as Fig.~\ref{fig:protocol}c of the main. Table~\ref{tab:deep_acc} reports the corresponding accuracies. LoRA + SDGM reaches the highest final average accuracy, on par with standard + SDGM, and the smallest backward transfer, at the price of a slightly lower plasticity than standard fine-tuning, since fewer channels remain available for the new tasks.

Fig.~\ref{fig:deep_share}a shows that the share of the update falling on the channels important for Task 1 stays at chance level for vanilla LoRA at every rank. As predicted by the theory, the low-rank constraint does not steer the update away from the units that carry Task 1, whereas SDGM sets this share to zero by construction. Fig.~\ref{fig:deep_share}b follows the Task 1 representation along the sequence. Standard fine-tuning moves it the most, LoRA less and LoRA + SDGM the least. Without a mask, the drift accumulates with each new task; adding a mask markedly slows it after the first tasks.

The benchmark departs from the assumptions of the theory in several ways: depth, convolutional layers, minibatch training over several epochs, and natural images. Two effects are absent from the two-layer model. First, the Task 1 representation still drifts under SDGM. In the theory, a frozen hidden unit reads the input data directly, so its output cannot change. In a deep network, by contrast, a frozen channel reads the outputs of the previous layer, part of which remain plastic, so its output can change even though its own weights do not. We expect this to be the main source of the residual drift. Second, since new channels are protected after each task, the plastic budget shrinks along the sequence, which eventually limits how many tasks can be learned with a fixed network. Finally, with three seeds and a single architecture, we claim the ordering of the methods rather than the precise size of the gaps.

\section{Trying the various procedures on a real dataset with the same architectural assumptions}
\label{app:MNIST}
To validate the predictions of our theory, we apply the proposed procedures to a sequence of simple tasks constructed from the MNIST dataset~\citep{726791}. The first task is a binary classification problem in which digits below 5 are assigned to class 0, while digits greater than or equal to 5 are assigned to class 1. The second task uses a different partition of the same dataset, with even digits assigned to class 0 and odd digits to class 1.

\begin{figure}[h!]
    \centering
    \includegraphics[width=0.8\linewidth]{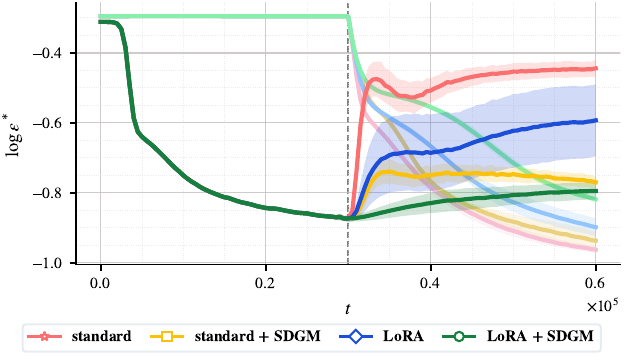}
    \caption{\textbf{Validation of the prediction on the MNIST dataset.} Generalization error dynamics on Task 1 and Task 2 when trained on a sequential MNIST task. The curves are averaged over 10 independent training realizations. The overall forgetting hierarchy is reproduced. Parameters: $N=784$, $K=10$, $L=5$.}
    \label{fig:app_MNIST}
\end{figure}

We train the model in an online learning setting using the full MNIST dataset. For each digit, the available examples are divided between the two tasks, resulting in $\mu=3\times10^4$ training examples per task. The $28\times28$ images are flattened into vectors of dimension $N=784$. The generalization error curves reported in Fig.~\ref{fig:app_MNIST} are averaged over 10 independent training runs, with variability arising from both the data split and the initialization of the student network modules.

The same qualitative behavior as in the theoretical setting is observed: forgetting is largest for the standard procedure and smallest when SDGM is applied to LoRA. The slowdown induced by the LoRA parameterization at the beginning of Task 2 training is also observed in the real-data experiments. The hyper-parameters are equal to those used for theoretical simulations, present in Appendix~\ref{app:hyperparameters}, the only modification being $N=784$ in order to match the input size.

\section{Hyper-parameters for numerical experiments}
\label{app:hyperparameters}
In this section, we summarize the hyper-parameters that were used to perform all numerical simulations. 
\begin{itemize}
    \setlength\itemsep{1em}
    \item Input dimension: $N=10^3$,
    \item Student hidden dimension: $K=10$,
    \item Teacher(s) hidden dimension: $M=5$,
    \item LoRA rank: $L=5$ (unless otherwise stated, e.g. Figure~\ref{fig:forgetting+transfer} and Section~\ref{sec:fixed_kappa}),
    \item $\mathrm{teacher}^\dagger-\mathrm{teacher}^\ddagger$ correlation coefficient: $c=0.5$ (unless otherwise stated),
    \item $\mathcal{S}_{\mathrm{frozen}}$ cardinality: $\kappa=K-L$ for LoRA + SDGM and $\kappa = 5$ for standard + SDGM (where $L = K$), except in Appendix~\ref{sec:fixed_kappa}, where $\kappa \in \{2, 5, 8\}$ for all $L$,
    \item LoRA prefactor: $\gamma=1$ for LoRA and LoRA + SDGM; $\gamma = \sqrt{K}$ with $L = K$ for standard + SDGM, written in the LoRA parameterization (Appendix~\ref{sec:SDGM_implementation}),
    \item Time horizon for each task: $\alpha=50$ (for the specialized case in Appendix~\ref{app:specialization} and Figure~\ref{fig:fixed_kappa}b, $\alpha=2000$ for Task 1 and $\alpha=8000$ for Task 2),
    \item Learning rates: $\eta_\mJ = \eta_\mB = \eta_\mA = \eta_\vh = 0.5$,
    \item Integration step for discretized ODEs using Euler's method: $h_{\mathrm{step}} = 0.05$.
\end{itemize}

The initialization for teachers and student networks are the following:
\begin{itemize}
    \setlength\itemsep{1em}
    \item Student first-layer weight $\mJ$: $\mJ_{ij}\sim\mathcal{N}(0,10^{-6})$ (except in Appendix~\ref{app:specialization}, where the initialization is described),
    \item Student readouts (for both tasks): $\vh^*_i\sim\mathcal{N}(0,10^{-4})$ (except in Appendix~\ref{app:specialization}),
    \item LoRA up-projection adapter $\mB$: see Appendix~\ref{app:LoRA} for classical LoRA framework or Equation~(\ref{eq:sdgm_mask}) for LoRA + SDGM,
    \item LoRA down-projection adapter $\mA$: $\mA=0$,
    \item $\mathrm{teacher}^\dagger$ first-layer weight $\mW^\dagger$: $\mW^\dagger_{ij}\sim\mathcal{N}(0,1)$,
    \item $\mathrm{teacher}^\ddagger$ first-layer weight $\mW^\ddagger$: $\mW^\ddagger = c\,\mW^\dagger+\sqrt{1-c^2}\,\mathbf{Z},\quad\mathbf{Z}_{ij}\sim\mathcal{N}(0,1)$, with $\mathbf{Z}$ independent of $\mW^\dagger$,
    \item Teacher 1 readouts: $\vv^\dagger_i = +1+n^\dagger_i,\qquad n^\dagger_i \sim \mathcal{N}(0,10^{-4})$,
    \item Teacher 2 readouts: $\vv^\ddagger_i = -1+n^\ddagger_i,\qquad n^\ddagger_i \sim \mathcal{N}(0,10^{-4})$.
\end{itemize}

\end{document}